%% file: main.tex
\documentclass{article}

\usepackage{arxiv}

\usepackage[utf8]{inputenc} 
\usepackage[T1]{fontenc}    
\usepackage{hyperref}       
\usepackage{url}            
\usepackage{booktabs}       
\usepackage{amsfonts}       
\usepackage{nicefrac}       
\usepackage{microtype}      
\usepackage{amsmath}
\usepackage{cleveref}       
\usepackage{lipsum}         
\usepackage{graphicx}
\usepackage[numbers]{natbib}
\usepackage{doi}
\usepackage{enumitem,kantlipsum}

\usepackage{xcolor}
\definecolor{LB}{rgb}{0.7, 0.85, 1.0}

\newcommand{\spot}{\texttt{SPOT}}
\newcommand{\cp}[1]{\tilde{\mathbf{x}}_{#1}}

\usepackage{todonotes}

\usepackage{enumitem}
\usepackage{subcaption} 
\usepackage[linesnumbered, ruled, vlined]{algorithm2e}
\usepackage{multirow}
\usepackage{enumerate}

\title{SPOT: \underline{S}patio-temporal  \underline{P}attern Mining and \underline{O}ptimization for Load Consolidation in Freight \underline{T}ransportation Networks}

\author{ 
    Sikai Cheng \\
    AI4OPT, Georgia Institute of Technology \\
    Atlanta, GA, USA \\
    \texttt{sikaicheng@gatech.edu} \\
    \And
    Amira Hijazi \\
    AI4OPT, Georgia Institute of Technology \\
    Atlanta, GA, USA \\
    \texttt{ahijazi6@gatech.edu} \\
    \And
    Jeren Konak \\
    AI4OPT, Georgia Institute of Technology \\
    Atlanta, GA, USA \\
    \texttt{ckonak3@gatech.edu} \\
    \And
    Alan Erera \\
    AI4OPT, Georgia Institute of Technology \\
    Atlanta, GA, USA \\
    \texttt{alan.erera@isye.gatech.edu} \\
    \And
    Pascal Van Hentenryck \\
    AI4OPT, Georgia Institute of Technology \\
    Atlanta, GA, USA \\
    \texttt{pvh@gatech.edu} \\
}

\renewcommand{\shorttitle}{SPOT}

\begin{document}

\maketitle

\input{abstract}

\keywords{Logistics, Clustering, Pattern Mining, Optimization, Load Consolidation}

\input{introduction}

\input{related-work}
\input{problem}
\input{method}

\input{experiments}
\input{experimental-result}
\input{conclusion}

\section*{Acknowledgment}

This work is partially supported by NSF Award 2112533 (NSF AI
Institute for Advances in Optimization).

\end{document}

%% file: abstract.tex
\begin{abstract}
Freight consolidation has significant potential to reduce
transportation costs and mitigate congestion and pollution. An
effective load consolidation plan relies on carefully chosen
consolidation points to ensure alignment with existing transportation
management processes, such as driver scheduling, personnel planning,
and terminal operations. This complexity represents a significant
challenge when searching for optimal consolidation
strategies. Traditional optimization-based methods provide exact
solutions, but their computational complexity makes them impractical
for large-scale instances and they fail to leverage historical
data. Machine learning-based approaches address these issues but often
ignore operational constraints, leading to infeasible consolidation
plans.

This work proposes \spot{}, an end-to-end approach that integrates the
benefits of machine learning (ML) and optimization for load
consolidation. The ML component plays a key role in the planning phase
by identifying the consolidation points through spatio-temporal
clustering and constrained frequent itemset mining, while the
optimization selects the most cost-effective feasible consolidation
routes for a given operational day.  Extensive experiments conducted
on industrial load data demonstrate that \spot{} significantly reduces
travel distance and transportation costs (by about 50\% on large
terminals) compared to the existing industry-standard load planning
strategy and a neighborhood-based heuristic. Moreover, the ML
component provides valuable tactical-level insights by identifying
frequently recurring consolidation opportunities that guide proactive
planning. In addition, \spot{} is computationally efficient and can be
easily scaled to accommodate large transportation networks.

\end{abstract}

%% file: introduction.tex
\section{Introduction}
\label{sec: introduction}

Freight transportation has grown rapidly over the last decade and has
become one of the largest components of the economy, accounting for
about 9.0\% of the U.S. gross domestic product and 10.3\% of the
U.S. labor force \cite{transportation_statistics_2024}. Given the vast
scale of this industry, efficient planning and operations are highly
desirable, not only by logistics companies aiming at increasing profits
\cite{sebastian-2012, gras-2022}, but also at the societal level to
reduce transportation externalities such as congestion, pollution,
noise, and accidents \cite{sathaye-2006, kabadurmus-2020,
  itf-2003-delivering-goods}. Academic research
\cite{anoop_2020_multimodal, van-heeswijk-2016} and industry practice
\cite{sunsetpacific_2024_consolidation, piechnik_2021_innovative}
indicate that a potential strategy to increase profits and mitigate
externalities lies in better utilization of container capacity and
reduction of partial loads. Partial loads, i.e., shipments that do not fully
utilize a container, are often considered unavoidable but remain an
important tactic to improve the flexibility of the transportation
network and improve customer service satisfaction \cite{williams-2013,
  attar-2024}.

Load consolidation, rooted in the concept of grouping various
shipments, parcels, or products into a single batch \cite{tyan-2003},
is considered an effective strategy for reducing the number of partial
loads and is widely applied in various logistics contexts, including
rail, ground, sea, and air transportation \cite{aboutalib-2024,
  baykasoglu-2011, mesa-arango-2013, van-andel-2018, monsreal-2024};
supply chain network design \cite{oguntola-2023, kulkarni-2024,
  greening-2023, greening-2023-1}; and urban congestion challenges
\cite{ouadi-2020}. However, transportation networks are highly
intricate, requiring extensive coordination of human and material
resources, with decisions often decentralized across multiple
management units \cite{crainic1997planning}, which makes load
consolidation across multiple terminals a complicated task. {\em Due
  to the fact that different terminals have different planners who do
  not have visibility over other terminals in the transportation
  network, it becomes crucial to define consolidation points well in
  advance} to ensure alignment between terminals, as well as with
other transportation management components, such as driver scheduling,
personnel planning, and terminal operations. This complexity makes
determining optimal consolidation strategies particularly challenging.

Novel algorithms and techniques designed to address these complexities
have been the topic of significant research. Specifically, prior
research in \cite{abdelwahab1990freight, cetinkaya-2005,
  bookbinder-2002, higginson1994shipment, nussy-2022, cetinkaya-2003}
examines optimal dispatching rules for \textit{temporal
  consolidation}, where orders are intentionally held and shipped
together, either after a fixed time interval or once a threshold
volume is reached.  However, these studies typically focus on a single
transportation route or shipment path, and do not address the more
challenging problem that involves groups of loads or multiple
routes. Another research direction \cite{baykasoglu-2011,
  mesa-arango-2013, monsreal-2024, kay-2021} formulates load
consolidation across multiple origin–destination pairs as a multi-stop
pick-and-delivery vehicle routing problem with time windows
(m-PD-VRPTW), also known as \textit{vehicle consolidation}. This
approach can be viewed as an extension of the vehicle routing problem
(VRP), and both mixed-integer programming (MIP)
\cite{mesa-arango-2013, monsreal-2024, kay-2021} and heuristic methods
\cite{baykasoglu-2011, monsreal-2024, kay-2021} have been developed
based on existing VRP algorithms. Nevertheless, m-PD-VRPTW typically
involves a significant computational overhead. The algorithms have
been demonstrated primarily on small-scale instances, and are
difficult to scale to real-world problems. In addition,
\citet{greening-2023, greening-2023-1} consider \textit{terminal
  consolidation} for middle-mile logistics network design, in which
loads are routed through predetermined intermediate terminals together
for consolidation. These models focus more on the long-term impact of
network arrangement and the coordination of other time
constraints. The integration of machine learning techniques, i.e.,
clustering, and association rule mining, is investigated in
\cite{aboutalib-2024, van-andel-2018} to evaluate the consolidation
performance.

Although considerable progress has been made in load consolidation,
existing methods still suffer from at least one of the following
challenges:

\textit{\textbf{(C1) - Computational Complexity.}}  The first
challenge concerns the high computational complexity inherent to load
consolidation problems when multiple origins and destinations are
involved. It is well-known that these problems are NP-hard, making it
impractical to solve large-scale instances optimally within a
reasonable time \cite{baykasoglu-2011, mesa-arango-2013,
  monsreal-2024, kay-2021, greening-2023}. Although heuristic
algorithms are often employed to mitigate the computational burden of
the exact solution methods, their performance can be hampered by the
vast search space associated with large-scale instances
\cite{monsreal-2024, kay-2021, greening-2023}.

\textit{\textbf{(C2) - Restricted Conditions for Operational Consolidation}} 
A second challenge lies in the restricted
conditions for load consolidation in practice, which are typically
addressed in the industry through long-term interactions between the
decentralized terminal planning units and other components within
transportation management frameworks \cite{camur-2024,
  serrano-2021}. A common issue arises when certain terminals cannot
accommodate consolidation due to a lack of necessary loading/unloading
equipment, space, or personnel. Additionally, certain routes may not
support load consolidation because of insufficient transportation
frequency or conflicts with current driver schedules. This operational
challenge is largely overlooked in the literature. Most proposed
algorithms are executed in a greedy and myopic manner
\cite{baykasoglu-2011, mesa-arango-2013, monsreal-2024, kay-2021},
assuming that the resulting consolidation routes are operational and
ignoring the long-term impact of load consolidation on the broader
transportation framework.

\textit{\textbf{(C3): Insufficient Precision in Consolidation Decisions.}} Although \citet{van-andel-2018} demonstrates that
consolidation opportunities exist within load clusters identified
based on latitude, longitude, and specific airports or ports,
recognizing these opportunities does not guarantee that loads can
actually be merged. On any given operational day, factors such as
departure times, package sizes, and vehicle capacities play a critical
role in determining whether loads can be consolidated. Similarly,
\citet{aboutalib-2024} highlight the concept of ``associated'' loads
from the pattern mining results, but also ignores the process of
constructing feasible consolidation decisions, which involves time and
capacity constraints, intermediate terminal selection, and optimal
vehicle routing.

\textit{\textbf{(C4) - Limited Use of Historical Data}} A fourth
challenge lies in the limited integration of historical data into
current consolidation models. Existing approaches
treat each problem instance as completely new, without leveraging past
instances and corresponding solutions to enhance efficiency. In
reality, logistics systems exhibit spatial and temporal patterns, such
as traffic hot spots \cite{bhattacharya-2013}, periodic or predictable
demand \cite{niu-2018, moscosolopez-2016, lee-2016}, and recurring
trip patterns \cite{pani-2014}. The spatial and temporal patterns
provide valuable insights into various aspects of logistics, including
inventory management \cite{stefanovic-2015}, monitoring and anomaly
detection \cite{hassan-2019}, and delay forecasting and management
\cite{pani-2014}. However, optimization and heuristic models for load
consolidation do not take advantage of these historical data insights.

\textit{\textbf{(C5) - Lack of Comprehensive Testing Datasets}} The
last challenge pertains to the shortage of large-scale, realistic
testing datasets in the field. For instance, \citet{baykasoglu-2011}
adapt a vehicle routing problem with time windows dataset to evaluate
their proposed algorithm, while \citet{mesa-arango-2013} rely on a
randomly generated network with only five nodes. Although
\citet{monsreal-2024} examine heuristic solutions, the largest
instance considered involves 58 clients, which does not reflect
real-world operational complexity. Without high-quality, large-scale
datasets that mirror actual operational environments, it is difficult
to validate the robustness and computational efficiency of proposed
algorithms.

\begin{figure*}[!tb]
    \centering
    \includegraphics[width=\textwidth]{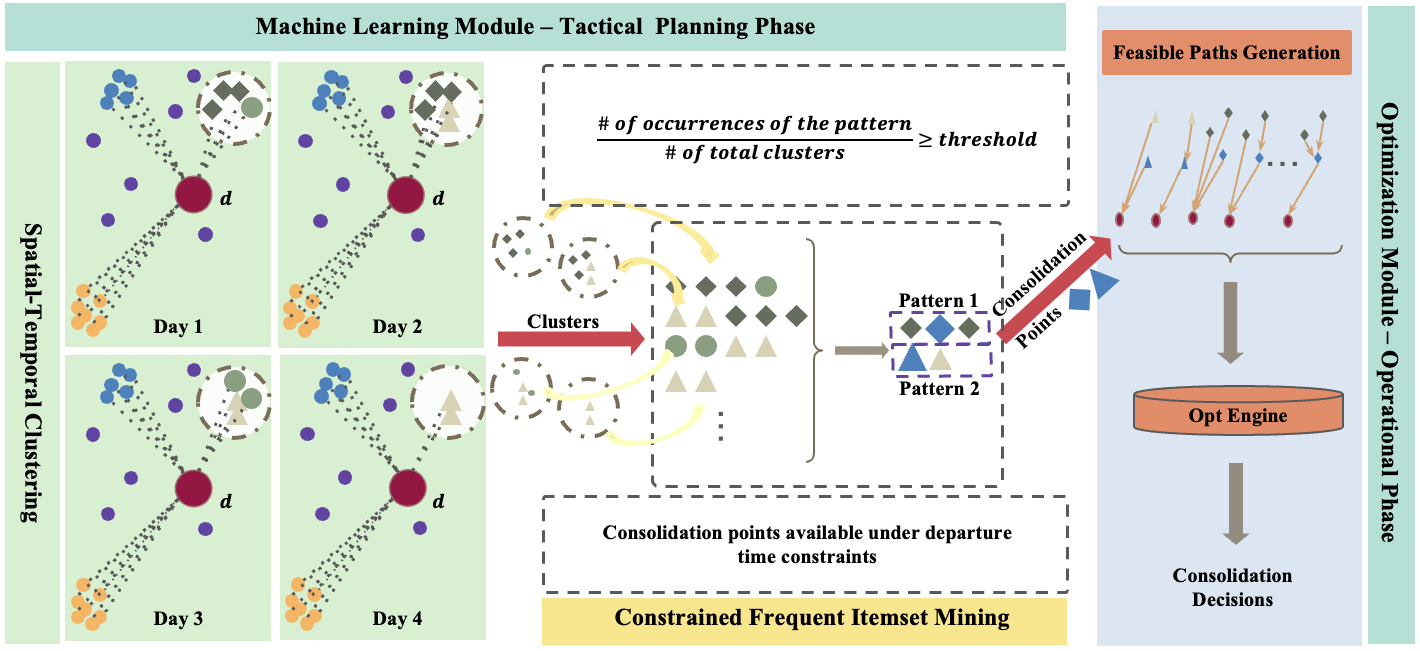}
    \caption{The Overview of \spot{}.}
    \label{fig: overview}
\end{figure*}

{\em This paper proposes \spot{}, an integrated end-to-end framework
  for load consolidation that combines Machine Learning (ML) for
  tactical planning and optimization for real-time operations (see
  Figure \ref{fig: overview}) to address these challenges}. During
tactical planning, due the decentralized nature of the considered
transportation networks, \spot{} identifies promising {\em
  consolidation points}, i.e., locations where load consolidations can
take place. \spot{} uses spatio-temporal clustering \cite{ansari-2019,
  georgoulas-2013, han-2015, izakian-2012, husch-2018} on partial
loads that share the same destination (left of Figure \ref{fig:
  overview}) and then frequent itemset pattern mining
\cite{agrawal1994fast, djenouri-2019, deng-2015} to analyze these
clusters, identify load groups that frequently appear together in the
historical data, and select potential consolidation candidates (middle
of Figure \ref{fig: overview}). During real-time operations, \spot{}
uses these consolidation points inside an optimization model to
determine consolidation decisions, using real-time data about
scheduled departure times, truck or container utilization, and costs
(right of Figure \ref{fig: overview}).

The contributions of \spot\ can be summarized as follows:
\begin{itemize}[leftmargin=.5cm]

\item \spot{} is the first integrated framework for real-world load
  consolidation tasks that integrates machine learning and
  optimization. \spot{} spans the entire process from extracting
  consolidation candidates from historical data for planning purposes
  to determining cost-saving and operationally feasible consolidation
  routes for a given operational day.

\item The ML module of \spot{} integrates spatio-temporal (ST)
  clustering with constrained frequent itemset mining (CFIM) to
  identify frequent consolidation candidates from historical data. In
  doing so, requirement \textit{\textbf{C4}} is effectively
  addressed. As highlighted before, isolating consolidation candidates
  in advance at the tactical level serves as a foundation for
  coordinating load consolidation between terminals planners and with
  other components of the transportation management systems, such as
  driver scheduling, personnel planning, and terminal management.


\item For operational consolidation decisions on a specific
  operational day, \spot{} uses an optimization model that utilize only the
  consolidation points identified by the ML module, where necessary
  preparations have been made in advance. The optimization is a
  mathematical programming model that determines the optimal
  consolidation route decisions within the context of terminal
  consolidation. By combining the ML output with the optimization
  model, the resulting consolidation decisions are both operationally
  feasible and effective, thereby addressing requirements
  \textbf{\textit{C2}} and \textbf{\textit{C3}}. In addition,
  these consolidation decisions are computed independently and in parallel
  for each destination. This makes it possible to determine
  consolidation decisions across the entire network efficiently
  addressing requirement \textit{\textbf{C1}}.

\item
  \spot{} is evaluated on real load data covering the entire
  U.S. transportation network, addressing requirement
  \textbf{\textit{C5}}. The experiments demonstrate the competitive
  performance of \spot{} in terms of travel distance reduction and
  cost savings, while also offering significant long-term insights.
\end{itemize}

\noindent
The rest of this paper is organized as follows. Section \ref{sec:
  related work} describes the related work. Section \ref{sec: problem}
introduces the problem considered in this paper. Section
\ref{sec:method} describes \spot{}, the
proposed load consolidation framework. Section \ref{sec: experiments}
describes the experimental setting and Section \ref{sec:
  experiment-result} describes the experimental results. Section
\ref{sec:conclusion} concludes the paper.

%% file: related-work.tex
\section{Related Work}
\label{sec: related work}

Load consolidation has been widely studied in both academia and industry. In addition to the literature on load consolidation presented in the introduction, the \spot{} approach is also closely related to research involving spatio-temporal (ST) clustering, frequent itemset mining (FIM), and optimization methods in logistics problems, as detailed in the following subsections.

\paragraph{Spatio-Temporal (ST) Clustering in Logistics}

The large volume of Spatio-Temporal (ST) data generated in recent
years \cite{belhadi-2020, ansari-2019}, along with developments of
geolocation technology (e.g., GPS), has led to a growing interest in
ST clustering techniques. These techniques group data points according
to latitude, longitude, and an extra time dimension
\cite{fanaee2012spatio, birant-2006}. Although ST clustering is
critically important in numerous domains, including image processing
and pattern recognition \cite{li-2004}, environmental studies
\cite{birant-2006}, traffic management \cite{anbaroglu-2014}, and
mobility data analysis \cite{bogorny-2010}, there are relatively few
studies exploring its potential in decision making for logistics. An
exception is \cite{qi-2011}, which proposes an efficient algorithm for
large-scale VRPTW by applying ST clustering to group
customers. \spot{} expands on this idea and leverages ST clustering
for load consolidation.

\paragraph{Frequent Itemset Mining (FIM) in Logistics}

Data mining has long been considered a key factor in the success of
logistics improvement initiatives \cite{frazzelle2002supply}, helping
to extract valuable insights from various areas, such as supply
activity profiles, transportation profiles, and warehouse activity
profiles. Specifically, focusing only on frequent itemset mining
(FIM), \citet{nohuddin-2017} proposes mining patterns of
cargo items frequently shipped to military camps, leading to an
ontology-like knowledge base for a specialized military transportation
network. In addition, \citet{lattner2008temporal} extend the
concept of an item to include events with temporal validity (e.g.,
truck travel, loading/unloading, terminal breakdown) to identify
frequent co-occurrences in historical data. These event patterns were
transformed into predictive rules, providing actionable insights. For
example, detecting co-occurring events with adverse outcomes enables
managers to take preventive measures. \citet{gutierrez2021data}
propose a robust and sustainable decision-making framework for urban
last-mile operations by mining patterns for products, customers,
zones, and drivers, and revealing significant factors that influence
decision-making accordingly. \spot{} relies on these insights and
treats partial loads in a transportation system as items; this makes
FIM a promising approach for pinpointing suitable consolidation
candidates from historical data.

\paragraph{Optimization in Logisitics}

Optimization models have been employed in nearly every aspect of
logistics to enhance efficiency, reduce costs, and improve
decision-making processes. For instance, \citet{chan-2006,
  jiang-2022} and \citet{ ye2024contextualstochasticoptimizationomnichannel}
propose optimization models for order fulfillment in multi-echelon
distribution networks and online retail networks.
\citet{cardenas-barron-2021} formulate an optimization model to
determine purchasing periods for oil, aiming at minimizing total
purchasing and inventory costs. Meanwhile, \citet{celik-2021}
address the storage replenishment routing problem using
mixed-integer programming. Recognizing the significant computational
overhead of such formulations, \citet{jiang-2022,
  cardenas-barron-2021} and \citet{celik-2021} also employ techniques such as
variable neighborhood search, MIP-based approximation heuristics and
routing-based heuristics to effectively address large-scale instances.

%% file: problem.tex
\section{Problem Statement}
\label{sec: problem}

This section specifies the load consolidation problem and the
notations used throughout this paper.

\paragraph{Freight Transportation Networks}

This paper considers a freight transportation network characterized by
spatial structure and temporal attributes and represented by a
directed graph $G = (V, A)$. The spatial structure comprises
approximately 1,000 {\em terminals} distributed across the United
States.  The temporal structure is organized around daily sorting
periods: each operational day is divided into sorting periods
(referred hereafter as \textit{sorts}), typically three to four hours
long, during which packages within the loads are
processed~\cite{bruys2024confidence}. Consequently, the inclusion of
sorts is crucial for accurately describing the load transportation
activities which take place between an origin terminal-sort pair and a
destination terminal-sort pair. In this setting, the set of nodes $V$
consists of terminal-sort pairs, and $A$ denotes the set of existing
direct routes among these nodes. Formally, each node $v = (v^\sigma,
v^\eta)$ is defined by its terminal $v^\sigma$ and sort
$v^\eta$. Furthermore, a load arriving at sort $v^\eta$ must adhere to
the latest arrival time to ensure timely processing; its departure
time from can be ready $v^\eta$ cannot be earlier that departure time
associated with the sort. These latest arrival time and earliest
departure time for sort $v$ are denoted by ${arr}(v^\eta)$ and
${dep}(v^\eta)$, respectively.

\paragraph{Load Consolidation}

A load $l$ in the transportation network is characterized by spatial
and temporal attributes as follows:
\begin{equation}
    l = \left( o_l = (o^\sigma_l, o^\eta_l), d_l = (d^\sigma_l, d^\eta_l), t_l, {due}_l \right),
\end{equation}
where $o_l$ represents its spatio-temporal origin node, $d_l$ its
spatio-temporal destination node, $t_l$ is its scheduled departure
time from $o_l$ ($t_l\geq {dep}(o^\eta_l))$, and ${due}_l$ is its due
date $d_l$ to ensure service. The number of transit days $\omega_l = {due}_l -
{day}(t_l)$ is the differences (in days) between the due date
and scheduled departure time.

\textbf{A consolidation point} is a node (load) where multiple loads
can be consolidated before traveling together across the network to
the common destination. These consolidation points must be identified
during tactical planning to synchronize the various terminals in the
network that operate largely independently. A \textbf{consolidated path}
can then be characterized as 
\begin{equation}
l = \left( o_l = (o^\sigma_l, o^\eta_l), h_l = (h^\sigma_l, h^\eta_l), d_l = (d^\sigma_l, d^\eta_l), t_l, t_h, {due}_l \right),
    \label{eq: consolidated routes}
\end{equation}
where $h_l \in V$ is a consolidation point and $t_h$ is the departure
time from $h_l$ after its consolidation.  This paper makes three
assumptions for the load consolidation framework:

\textit{\textit{(A1) Partial Loads Only:}} Loads that are already
fully utilized are excluded from consolidation, as there is no clear
incentive to split or reconfigure a fully utilized load. By contrast,
combining multiple partial loads into fewer trailers can substantially
reduce the total number of trips and thus lower transportation costs.

\begin{figure}[!t]
    \centering
    \begin{subfigure}{0.6\linewidth}
        \centering
        \includegraphics[width=\textwidth]{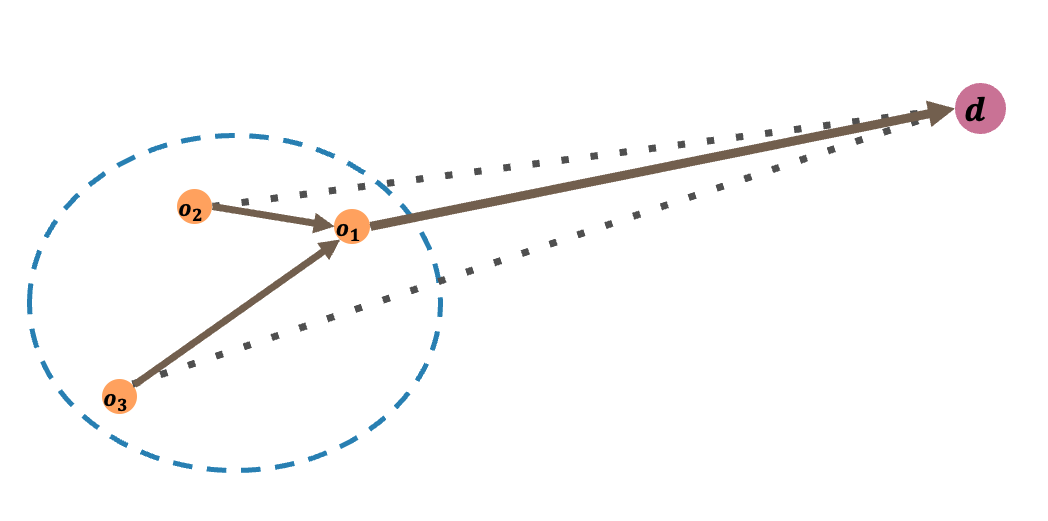}
        \caption{Consolidating Loads for a Single Destination.}
        \label{fig: consolidation_single_destination}
    \end{subfigure}
    \begin{subfigure}{0.6\linewidth}
        \centering
        \includegraphics[width=\textwidth]{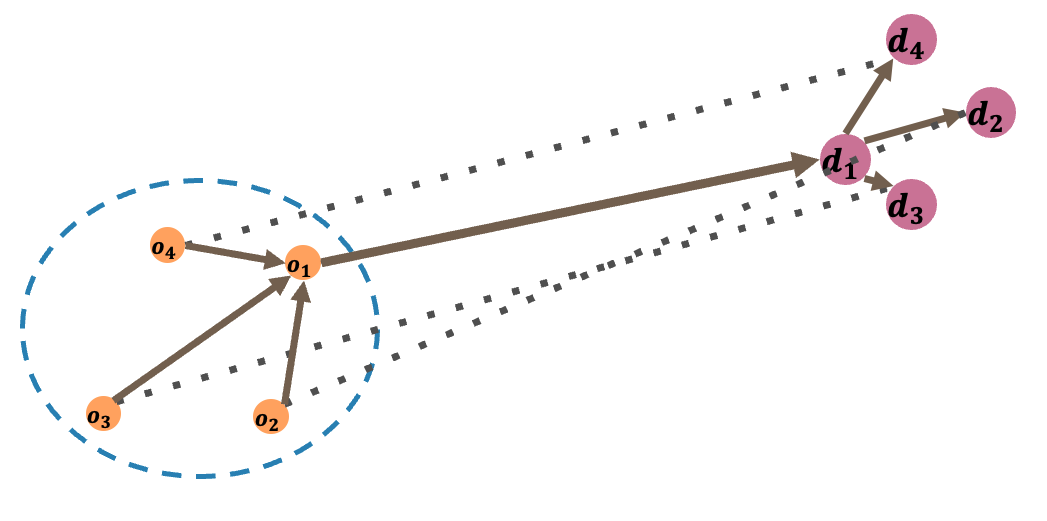}
        \caption{Consolidating Loads for Multiple Destinations.}
        \label{fig: consolidation_multiple_destination}
    \end{subfigure}
    \caption{Illustrating Load Consolidations.}
    \label{fig: consolidation single or multiple destinations}
\end{figure}

\textit{\textit{(A2) Same Due Date, and Destination:}} Only loads
sharing the same destination ($d_l$) and due date (${due}_l$) can be
consolidated (hereafter referred to as ``consolidation condition''),
as illustrated in the Figure.~\ref{fig:
  consolidation_single_destination}. The consolidation of loads with
multiple destinations that are in close proximity (as shown in
Figure.~\ref{fig: consolidation_multiple_destination}) is a natural
extension of \spot{}; it is beyond the scope of this work for
ease of deployment reasons.

\textit{\textit{(A3) Consolidation at Existing Origins:}} The choice of
consolidation points is restricted to the existing origins of the
considered partial loads. In Equation \eqref{eq: consolidated routes},
this means that $h_l$ must already be the origin of another load
$\mathfrak{h}$ and that both $l$ and $\mathfrak{h}$ are effectively
consolidated together. Again, this assumption is motivated by practical
deployment reasons. 




    

The goal of load consolidation is to find a feasible and cost-effective consolidation plan that
\begin{itemize}[leftmargin=.5cm]
\item defines the set of potential consolidation points $H \subset V$
  at the technical planning level, using historical records to ensure
  alignment with driver scheduling, personnel planning, and terminal
  operations.

\item chooses consolidation paths for partial loads using the
  predefined set of consolidation points to minimize the total
  transportation cost, subject to constraints on scheduled departure
  times, planned volume, and trailer capacity.
\end{itemize}

%% file: method.tex
\section{The \spot{} Framework for Load Consolidation}
\label{sec:method}

This section describes \spot{}, an end-to-end framework that uses
machine learning at the tactical planning level and optimization
during operations, as shown in Figure~\ref{fig: overview}. This
section reviews the details of these components and how they
interact. Since \spot{} is implemented independently for each
destination, the discussion below focuses on a single
destination. Note that the decomposition by destination has
significant scalability benefits. 

\subsection{The Machine Learning Component}

The goal of the ML component is to identify potential consolidation
points for destination $d$ through spatio-temporal clustering and
constrained frequent itemset mining. 

\input{method-st-clustering}
\input{method-cfim}
\input{method-st-clustering-vis}
\input{method-cfim-vis}
\input{method-opt}

\input{method-opt-vis}

%% file: method-st-clustering.tex
\subsubsection{Spatio-Temporal Clustering (ST Clustering)}\label{sec: st clustering}

For a given destination $d$, the ST clustering receives as input all
partial loads in the historical dataset with $d_l = d$. The ST
clustering then groups partial loads that are ``close'' to one another
and share the same due date (${due}_l$). For each partial load $l$, an
event data point $\mathbf{x}_l$ is defined to capture the key
attributes for clustering:
\begin{equation}
    \mathbf{x}_l = (o_l, {due}_l)
    \label{eq: spatio-temporal nodes}
\end{equation}
where
\begin{itemize}
    \item the spatial component, $\mathbf{x}_l(s) = o^\sigma_l$, represents the load origin;
    \item the temporal component, $\mathbf{x}_l(t) = (o^\eta_l, {due}_l)$, captures the origin sort and the load due date.
\end{itemize} 

\noindent
Partial loads are clustered based on their spatial and temporal
proximity. While latitude–longitude coordinates (along with Euclidean
or Haversine distances) and the absolute difference between timestamps
are commonly used for ST clustering \cite{hudjimartsu-2017,
di-martino-2017, georgoulas-2013, han-2015, ansari-2019}, they can be
misleading in the long-haul load transport context, where
consolidation may occur along the route. For example, consider three
partial loads $l_1$ $l_2$, and $l_3$ destined for $d$ as shown in
Figure~\ref{fig: illustrate the polar coordinate}. A conventional
metric based on $Dist(\mathbf{x_1(s)}, \mathbf{x_2(s)})$ and $\vert
(\mathbf{x_1(t)} - \mathbf{x_2(t)}) \vert$ would conclude that the
distance between $l_1$ and $l_2$ is too large, thus excluding them
from the same cluster. However, $o_1^\sigma$ lies on the route from
$o_2^\sigma$ to $d$, and thus a modest detour would enable $l_2$ to
consolidate with $l_1$ if it reaches $o_1^\sigma$ before $t_1$.

{\em Polar coordinates} capture such potential consolidations. Let
$\varphi(\mathbf{x})$ denote the angle between the origin-destination
route ($\mathbf{x}(s)$,$d$) and a reference direction (e.g., West)
quantifying the orientation of the load route relative to $d$. The
spatial proximity between two nodes can then be defined as:
\begin{equation}
    D_s ( \mathbf{x}_1, \mathbf{x}_2  ) = \vert \varphi(\mathbf{x}_1) - \varphi(\mathbf{x}_2)  \vert.
    \label{eq: D_s}
\end{equation}
which measures the difference in route alignment. This formulation
ensures that loads with similar directional orientations can be
clustered together.

\begin{figure}[!t]
    \centering \includegraphics[width=0.6\linewidth]{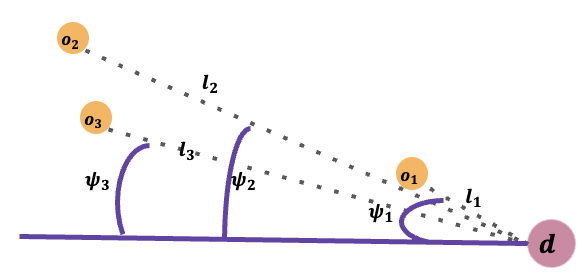} \caption{Consolidation
    Along the Origin-Destination Path.}  \label{fig: illustrate the
    polar coordinate}
\end{figure}

The temporal proximity metric is defined as the difference between the due dates of the partial loads:
\begin{equation}
    D_t ( \mathbf{x}_1, \mathbf{x}_2  ) = \vert {due}_1 - {due}_2 \vert.
    \label{eq: D_t}
\end{equation}

Given the spatio-temporal proximity measures defined above, \spot{}
uses DBSCAN (Density-Based Spatial Clustering of Applications with
Noise) \cite{schubert-2017} to cluster partial loads. DBSCAN is
well-suited for this task because consolidation corresponds to
identifying high-density regions where loads exhibit similar spatial
and temporal characteristics. Unlike partitioning methods such as
k-means \cite{macqueen1967} or Partitioning Around Medoids
(PAM) \cite{kaufman-1990}, which require specifying a fixed number of
clusters, or hierarchical methods such as BIRCH \cite{zhang-1996} or
Chameleon \cite{karypis-1999} that focus on nested structures, DBSCAN
can detect clusters of arbitrary shape and size based on the defined
proximity criteria. Specifically, spatio-temporal nodes are grouped in
one cluster if they satisfy the following conditions:
\begin{equation}
    D_s ( \mathbf{x}_1, \mathbf{x}_2  ) \leq \epsilon \, \wedge \,  D_t ( \mathbf{x}_1, \mathbf{x}_2  ) = 0,
    \label{eq: st clustering}
\end{equation}
where $\epsilon$ is a pre-defined angle alignment threshold. By enforcing the temporal condition, only loads with the same due date are eligible for consolidation.  DBSCAN takes, as input, the set of event data points
\begin{equation}
    \mathbf{X} = \{\mathbf{x}_1, \mathbf{x}_2, ... \mathbf{x}_{\vert\mathbf{X}\vert} \}
\end{equation}
which represents all partial loads destined to $d$, and outputs a set of clusters 
\begin{equation}
    \mathcal{C}= \{ C_1, C_2, ..., C_{N} \}
\end{equation}
where each $C_i$ is a subset of $\mathbf{X}$. 

%% file: method-cfim.tex
\subsubsection{Constrained Frequent Itemset Mining}
\label{sec: CFIM}

Once the clusters are identified, \spot{} uses Constrained Frequent
Itemset Mining (CFIM) to identify loads that are frequently
co-occurring for each day of the week (abbreviated as ``dow'').  This
will make it possible to select the consolidation points, which is the
ultimate goal of CFIM. As shown in Figure~\ref{fig: overview}, CFIM
takes, as input, the set of clusters $C$ for destination $d$. Each
cluster $C_i$ contains a set of partial loads, with each partial load
$l$ represented by event data point $\mathbf{x_l}$.  CFIM does not use
the $\mathbf{x_l}$ data points directly, since they are specific to
specific days and the goal is to extract repeating patterns. Instead,
it uses data points of the form $\tilde{\mathbf{x}}_l$, where the
exact due date has been replaced by meaningful features, i.e., its
corresponding day-of-week and the number of transit days. The clusters
remain the same, but the data has been abstracted to enable the
identification of frequent consolidation patterns on a day-of-week
basis. The updated representation is defined as
\begin{equation}
    \begin{aligned}
        \tilde{\mathbf{x}}_l & = (o_l, {due}_l^{\textit{dow}}, \omega_l) \\
        \mathcal{X} & = \{\tilde{\mathbf{x}}_1, \tilde{\mathbf{x}}_2,  ... \tilde{\mathbf{x}}_M\} \\
        \tilde{C}_k & = \{ \tilde{\mathbf{x}}_l \ \mid \ \mathbf{x} \in C_k \} \\
        \tilde{\mathcal{C}} & = \{ \tilde{C}_1, \ldots, \tilde{C}_{N} \}.
    \end{aligned}
\end{equation}
Intuitively, the goal is to find elements of $\mathcal{X}$ that
frequently occur together in the same clusters
in $\tilde{\mathcal{C}}$.

Following \cite{agrawal1994fast}, a \textit{consolidation candidate}
(itemset) $S \subset \mathcal{X}$ is deemed \textit{frequent} if its
support, i.e., the fraction of clusters containing $S$, meets or
exceeds a pre-defined threshold \textit{min\_sup}, where the
support of a set $S$ is given by
\begin{equation}
    sup(S) = \frac{ \sum_{k=1}^{\vert \tilde{\mathcal{C}} \vert} \mathbf{1}_{\left( S \subseteq \tilde{C}_k \right)}}{\vert \tilde{\mathcal{C}} \vert}
    \label{eq: support of an itemset}
\end{equation}
Any frequent $S$ discovered in this manner indicates that its constituent loads frequently co-occur for $d$.

The consolidation candidates so identified are not guaranteed to
contain actual consolidation due to temporal constraints. To remedy
this limitation, \spot{} each consolidation candidate to include at
least one time-feasible consolidation opportunity. This
time-feasibility check filters out infeasible consolidation
candidates, thereby retaining only the most useful information for
tactical planning and reducing the search space more effectively for
the subsequent optimization model. In particular, the time-feasibility
check, $\kappa$, for each candidate $S$ is as follows:
\begin{equation}
    \begin{aligned}
        \kappa (S) =  \exists & \ \tilde{\mathbf{x}}_i, \tilde{\mathbf{x}}_j \in S:  \\
        &  {dep}(o_i^\eta)+ \tau (o_i^\sigma, o_j^\sigma)  \leq 
         {arr}(o_j^\eta) + ( \omega_i - \omega_j ) \; \vee  \\
        &  {dep}(o_j^\eta) + \tau (o_j^\sigma, o_i^\sigma)  \leq 
        {arr}(o_i^\eta) +  ( \omega_j - \omega_i ). \\
    \end{aligned}
    \label{eq: time feasible constraints}
\end{equation}
Here $o_i^\sigma$ and $o_j^\sigma$ denote the origins of loads $i$ and
$j$, respectively, and $\tau(\cdot, \cdot)$ is the traveling time
between two origins; the terms ${dep}(o^\eta)$ and ${arr}(o^\eta)$
represent the earliest departure time and latest arrival time of a
load with respect to sort $o^\eta$. The differences in transit days,
i.e., $( \omega_i - \omega_j)$ and $( \omega_j - \omega_i )$, ensure
that loads requiring longer transit times can still be consolidated
with those having fewer transit days, provided that their routes
overlap en route to the common destination. Observe that, whenever an
itemset $S$ satisfies $\kappa$, so does any superset of $S$, which
implies that $\kappa$ is a \textit{monotone} constraint. This property
is highly desirable in frequent itemset mining.

To efficiently extract these constrained frequent itemsets, \spot{}
uses the Frequent Pattern Growth (FP-growth)
algorithm \cite{pei-2002}. Compared to the Apriori
algorithm \cite{shankar-2016, bhandari-2015}, the FP-growth algorithm
is computationally more efficient \cite{han-2007} because it uses
a \textit{divide-and-conquer} strategy to mine a compressed FP-tree
representation of the dataset. In the constrained version of
FP-growth, the feasibility check is incorporated at each resulting
pattern of the FP-tree to discard infeasible candidate sets.

Let $\mathfrak{S}$ denote the output of FP-growth, which is the set of
consolidation candidates that satisfy the time-feasibility check. For
any $S \in \mathfrak{S}$, the corresponding \textit{consolidation
points} are defined as
\begin{equation}
    H(S) = \{ o_l | \tilde{\mathbf{x}}_l \in S,\ \exists\ \tilde{\mathbf{x}}_{l'} \in S:  \tilde{\mathbf{x}}_{l'}\neq \tilde{\mathbf{x}}_l  \, \wedge \, \kappa'(\tilde{\mathbf{x}}_{l'}, \tilde{\mathbf{x}}_l) \text{ is true} \}.
\end{equation}
where 
\begin{equation}
    \kappa'(\tilde{\mathbf{x}}_{l'}, \tilde{\mathbf{x}}_l) == ({dep}(o_{l'}^\eta)+ \tau (o_{l'}^\sigma, o_l^\sigma)  \leq  {arr}(o_l^\eta) + ( \omega_{l'} - \omega_l ) )
    \label{eq: kappa'}.
\end{equation}

Accordingly, the set of consolidation points for $d$ is then defined as 
\begin{equation}
    H = \cup_{S \in \mathfrak{S}}\ H(S)
    \label{eq: consolidation points}.
\end{equation}
Intuitively, a consolidation point is the origin of a load that can be
consolidated with at least one other load. For instance, $o_1$ in
Figure \ref{fig: illustrate the polar coordinate} is a consolidation
point.

%% file: method-st-clustering-vis.tex
\subsubsection{Ilustration of the Machine Learning Component}

\begin{figure}[!t]
    \centering
    \includegraphics[width=0.9\linewidth]{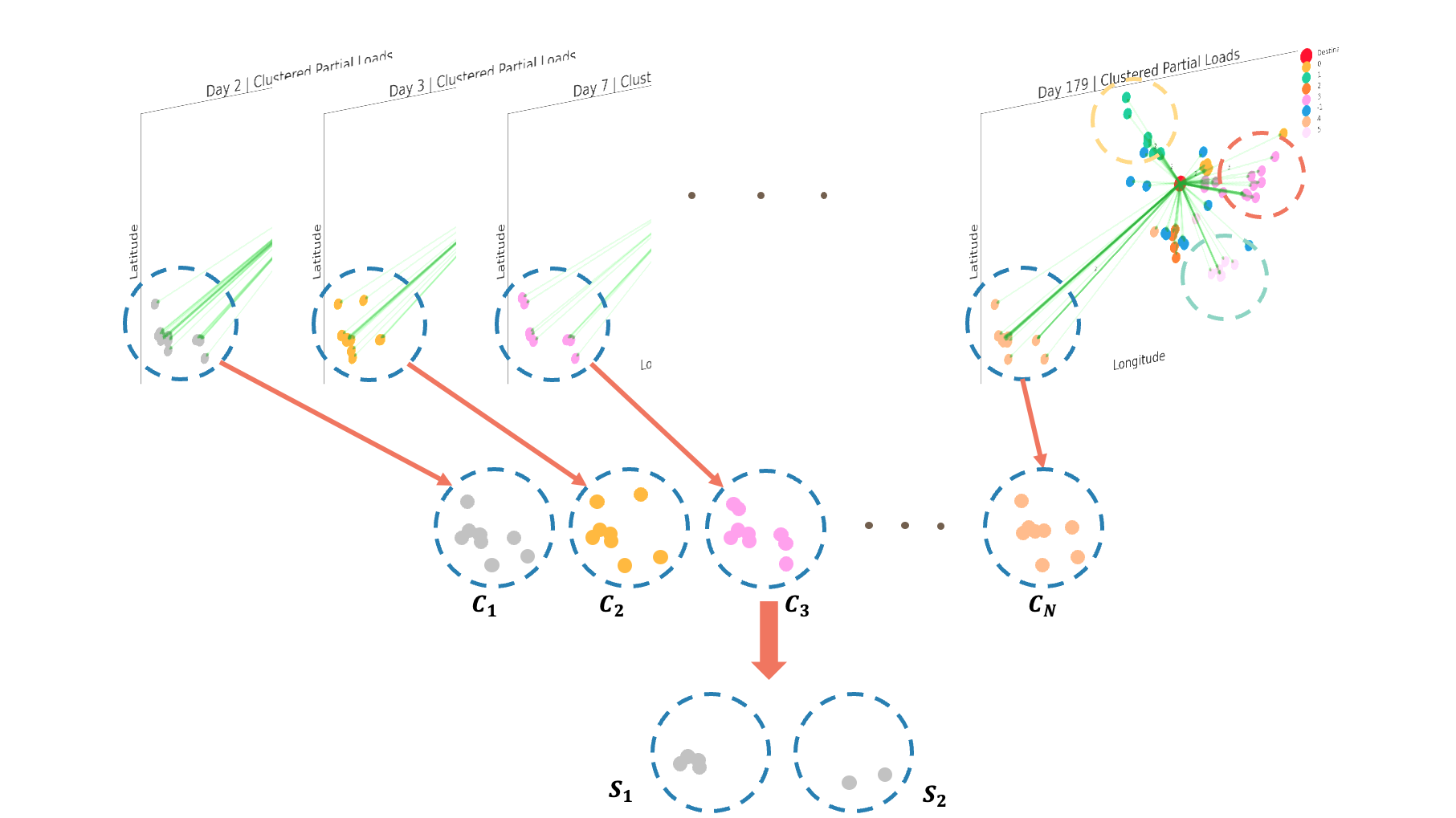}
    \caption{Illustration of the Machine Learning Component.}
    \label{fig: illustration of st clustering}
\end{figure}

Figure~\ref{fig: illustration of st clustering} offers a complete
illustration of the ML component, highlighting the interaction between
the clustering process and the subsequent CFIM. In the first phase,
the DBSCAN algorithm forms clusters, utilizing the spatial and
temporal distances $D_s$ and $D_t$ defined in \eqref{eq: D_s}
and \eqref{eq: D_t}. Moreover, since \eqref{eq: st clustering}
mandates that each cluster has the same due date, the clustering
results are automatically separable by due date. As depicted in the
figure, certain clusters emerge in similar locations and contain
overlapping data points. The second CFIM phase extracts recurring
patterns found across multiple clusters and including consolidation
points. In Figure~\ref{fig: illustration of st clustering}, pattern
$S_1$ appears in clusters $C_1$, $C_2$, and $C_3$, while $S_2$ is
present in clusters $C_1$, $C_2$, and $C_N$.  Both $S_1$ and $S_2$ are
recognized as frequent patterns. 

%% file: method-cfim-vis.tex
\begin{table}[!t]
  \centering
  \begin{subtable}[t]{0.35\textwidth}
    \centering
    \begin{tabular}{|c|l|}
    \hline
    ClusterID & Points \\
    \hline
    $\tilde{\mathcal{C}}_1$ & $\cp{7}$, $\cp{5}$, $\cp{8}$, $\cp{10}$ \\
    $\tilde{\mathcal{C}}_2$ & $\cp{10}$, $\cp{8}$, $\cp{5}$, $\cp{2}$, $\cp{9}$ \\
    $\tilde{\mathcal{C}}_3$ & $\cp{8}$, $\cp{3}$, $\cp{10}$, $\cp{6}$, $\cp{1}$, $\cp{4}$ \\
    $\tilde{\mathcal{C}}_4$ & $\cp{4}$, $\cp{5}$, $\cp{2}$, $\cp{9}$ \\
    $\tilde{\mathcal{C}}_5$ & $\cp{7}$, $\cp{6}$, $\cp{3}$ \\
    $\tilde{\mathcal{C}}_6$ & $\cp{3}$, $\cp{6}$, $\cp{5}$, $\cp{9}$ \\
    $\tilde{\mathcal{C}}_7$ & $\cp{1}$, $\cp{7}$, $\cp{10}$, $\cp{2}$, $\cp{8}$ \\
    \hline
    \end{tabular}
    \caption{Clusters as Input to CFIM.}
    \label{table: clusters as input}
  \end{subtable}
  \hspace{0.05\textwidth}
  \begin{subtable}[t]{0.3\textwidth}
    \centering
    \begin{tabular}{|c|l|}
    \hline
    Point & Consolidable Points \\
    \hline
    $\cp{1}$ & $\cp{3}$, $\cp{4}$, $\cp{6}$, $\cp{7}$, $\cp{8}$ \\
    $\cp{2}$ & $\cp{3}$, $\cp{4}$, $\cp{7}$, $\cp{9}$ \\
    $\cp{3}$ & $\cp{4}$, $\cp{5}$, $\cp{6}$ \\
    $\cp{4}$ & $\cp{7}$ \\
    $\cp{5}$ & $\cp{8}$, $\cp{9}$, $\cp{10}$ \\
    $\cp{6}$ & $\cp{9}$, $\cp{10}$ \\
    $\cp{7}$ & $\cp{9}$ \\
    $\cp{8}$ & --- \\
    $\cp{9}$ & --- \\
    $\cp{10}$ & --- \\
    \hline
    \end{tabular}
    \caption{Time-Feasible Consolidability between Points. As an example, $\cp{7}$ can consolidate at $\cp{9}$ but the opposite is not feasible.}
    \label{table: reachability between points}
  \end{subtable}
  \caption{Illustration of the CFIM Inputs.}
  \label{table: CFIM inputs}
\end{table}

\begin{table}[!t]
  \centering \begin{subtable}[t]{0.35\textwidth} \centering \begin{tabular}{|c|c||c|c|} \hline
  Points & Count & Points & Count \\ \hline $\cp{10}$ & 4 & $\cp{5}$ &
  4 \\ $\cp{8}$ & 4 & $\cp{2}$ & 3 \\ $\cp{3}$ & 3 & $\cp{6}$ & 3 \\
  $\cp{9}$ & 3 & $\cp{1}$ & 2 \\ $\cp{4}$ & 2 & $\cp{7}$ &
  2 \\ \hline \end{tabular} \caption{Appearance Count (Number of
  Clusters).}  \label{table: appearance count of
  points} \end{subtable} \hspace{0.05\textwidth} \begin{subtable}[t]{0.3\textwidth} \centering \begin{tabular}{|c|l|} \hline
  ClusterID & Points \\ \hline $\tilde{\mathcal{C}}_1$ & $\cp{10}$,
  $\cp{5}$, $\cp{8}$, $\cp{7}$ \\ $\tilde{\mathcal{C}}_2$ & $\cp{10}$,
  $\cp{5}$, $\cp{8}$, $\cp{2}$, $\cp{9}$ \\ $\tilde{\mathcal{C}}_3$ &
  $\cp{10}$, $\cp{8}$, $\cp{3}$, $\cp{6}$, $\cp{1}$, $\cp{4}$ \\
  $\tilde{\mathcal{C}}_4$ & $\cp{5}$, $\cp{2}$, $\cp{9}$, $\cp{4}$ \\
  $\tilde{\mathcal{C}}_5$ & $\cp{3}$, $\cp{6}$, $\cp{7}$ \\
  $\tilde{\mathcal{C}}_6$ & $\cp{5}$, $\cp{3}$, $\cp{6}$, $\cp{9}$ \\
  $\tilde{\mathcal{C}}_7$ & $\cp{10}$, $\cp{8}$, $\cp{2}$, $\cp{7}$,
  $\cp{1}$ \\ \hline \end{tabular} \caption{Reorganized Clusters
  Sorted by Appearance Count} \label{table: reorganized
  clusters} \end{subtable} \caption{Illustration of the FP-growth
  Preprocessing.}  \label{table: fp-growth preprocessing}
\end{table}

\begin{figure}[!t]
    \centering
    \begin{subfigure}{0.43\linewidth}
        \centering
        \includegraphics[width=\textwidth]{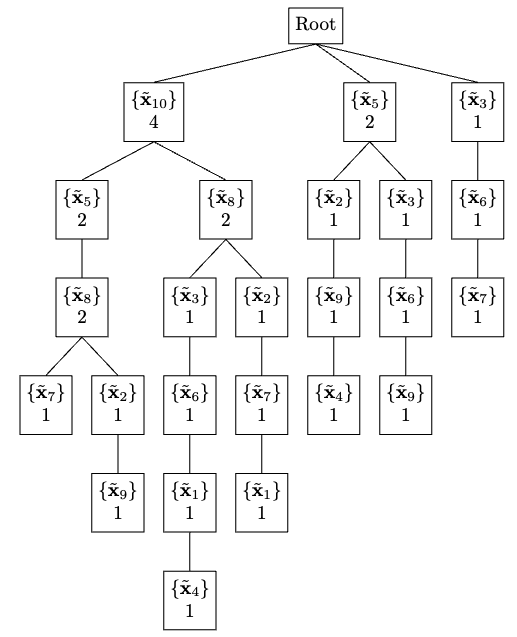}
        \caption{The Whole FP Tree.}
        \label{fig: CFIM - fp whole tree}
    \end{subfigure}
    \hfill
    \begin{subfigure}{0.26\linewidth}
        \centering
        \includegraphics[width=\textwidth]{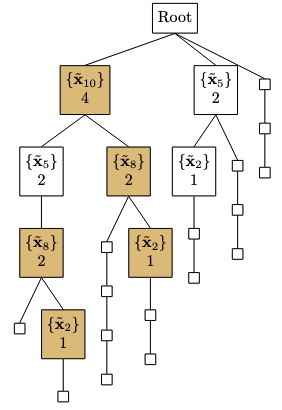}
        \caption{The conditional FP Tree - $\cp{2}$.}
        \label{fig: CFIM - fp subtree 2}
    \end{subfigure}
    \hfill
    \begin{subfigure}{0.26\linewidth}
        \centering
        \includegraphics[width=\textwidth]{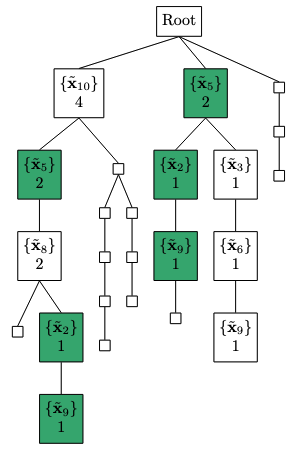}
        \caption{The conditional FP Tree - $\cp{9}$.}
        \label{fig: CFIM - fp subtree 9}
    \end{subfigure}
    \caption{Illustration of the FP-growth Algorithm.}
    \label{fig: an illustration of the fp-growth algorithm}
\end{figure}

Tables~\ref{table: CFIM inputs} and \ref{table: fp-growth
preprocessing}, and Figure~\ref{fig: an illustration of the fp-growth
algorithm} together present a complete example that illustrates how
the FP-growth algorithm is applied to the clusters generated during
the clustering phase. This process is used to extract frequently
co-occurring feasible consolidation candidates. Table~\ref{table: CFIM
inputs} consists of two parts:
\begin{itemize}[leftmargin=.5cm]

\item
Table~\ref{table: clusters as input} lists the input clusters, based
on a small-scale example with seven clusters and ten points.

\item
Table~\ref{table: reachability between points} presents the sorting
time constraints defined in \eqref{eq: kappa'}.
\end{itemize}

\noindent
The points are sorted by their earliest departure times, meaning that
points with smaller indices are more likely to be consolidated with
those that appear later. The reverse, however, is not feasible due to
the departure time constraints. Each cluster is then reordered based
on descending frequency of point occurrences in the datase. These
frequencies are provided in Table~\ref{table: appearance count of
points}. Table~\ref{table: reorganized clusters} shows the reorganized
clusters, now following a unified point order:
$\cp{10}, \cp{5}, \cp{8}, \cdots, \cp{7}$.  This reordering reduces
redundancy and promotes a more compact FP-tree structure by placing
high-frequency points near the root, thereby encouraging shared
prefixes across clusters. For example, in Table~\ref{table:
reorganized clusters}, $(\cp{10}, \cp{5})$ is shared by
$\tilde{\mathcal{C}}_1$ and $\tilde{\mathcal{C}}_2$, and
$(\cp{10}, \cp{8})$ appears in both $\tilde{\mathcal{C}}_3$ and
$\tilde{\mathcal{C}}_7$, each prefix occurring twice. By contrast, in
the original clusters from Table~\ref{table: clusters as input}, only
one shared prefix of length one, $(\cp{7})$, exists—shared between
$\tilde{\mathcal{C}}_1$ and $\tilde{\mathcal{C}}_5$.

Using the reorganized clusters, the FP-growth algorithm constructs the
full FP-tree shown in Figure~\ref{fig: CFIM - fp whole tree}. Each
path from root to leaf represents a cluster, and the number at the
leaf node indicates the frequency of that exact cluster. Here, since
all clusters are unique, each leaf has a count of 1. For example, the
leftmost path, $(\cp{10}, \cp{5}, \cp{8}, \cp{7})$, represents
$\tilde{\mathcal{C}}_1$, which appears only once. For internal nodes,
the number indicates how many clusters share that prefix path. In this
same subtree, the prefix $(\cp{10}, \cp{8})$ has count 2 is shared by
$\tilde{\mathcal{C}}_3$ and $\tilde{\mathcal{C}}_7$, but the longer
prefix $(\cp{10}, \cp{8}, \cp{2})$ has count 1 since it appears only
in $\tilde{\mathcal{C}}_7$.

Next, conditional FP-trees are derived from the full FP tree, focusing
on prefix paths that terminate at specific points. Figure~\ref{fig:
CFIM - fp subtree 2} shows the conditional FP-tree for $\cp{2}$, while
Figure~\ref{fig: CFIM - fp subtree 9} focuses on $\cp{9}$. If a
pattern is deemed as "frequent" when it appears in at least two of the
seven clusters, then $(\cp{10}, \cp{8}, \cp{2})$ is a frequent pattern
for $\cp{2}$, and $(\cp{5}, \cp{2}, \cp{9})$ is frequent for
$\cp{9}$. However, based on the sorting time constraints in
Table~\ref{table: reachability between points}, the first pattern does
not have any valid consolidation paths and is discarded. In contrast,
both $(\cp{2} \rightarrow \cp{9})$ and $(\cp{5} \rightarrow \cp{9})$
are feasible under sorting time constraints, making
$(\cp{5}, \cp{2}, \cp{9})$ a valid consolidation candidate, with
$\cp{9}$ serving as the consolidation point.

%% file: method-opt.tex
\subsection{The Optimization Component}


During real-time operations, there are $\mathfrak{L}$ planned loads
destined for $d$ on a specific day, among which $L$ are partial. The
set of consolidation candidates $\mathfrak{S}$ and consolidation
points $H$ is also available from the tactical planning stage.  The
optimization module selects the most cost-effective consolidation
decisions by leveraging real-time data on scheduled departure times,
truck or container utilization, and costs. It starts by identifying
feasible consolidation routes for each partial load using
$\mathfrak{S}$ and $H$. \spot{} uses a mathematical model to determine
the optimal consolidation routes.

\subsubsection{Feasible Path Generation}\label{sec: feasible path generation}

The feasible path generation only considers the subset $L_C$ of loads
with a valid consolidation load-pair, i.e.,
\begin{equation*}
    L_C = \{ l \in L\ |\ \exists\ S \in \mathfrak{S}, \text{ s.t. } \tilde{\mathbf{x}}_l \in S \},
\end{equation*}
The set $P_l$ of all feasible paths for a load $l \in L_C$ contains two types of
paths:
\begin{itemize}[leftmargin=.5cm]
    \item \textit{The Direct Route} $(o_l,d_l)$ from origin to destination;
    \item \textit{Consolidation Routes} of the form $(o_l, h, d_l)$, which includes a consolidation point $ h \in H$.
\end{itemize}
It is defined as
\begin{equation}
    \begin{aligned}
        & P_l =  \{ (o_l, d_l) \}\; \cup \; \{ (o_l, o_{\mathfrak{h}}, d_l)\ |\ \mathfrak{h} \in L_C\ \&\ o_{\mathfrak{h}} \in H\ \&\ d_{\mathfrak{h}} = d_l\ \&\ {due}_l = {due}_{\mathfrak{h}}\ \&\
      \kappa''(\tilde{\mathbf{x}}_l, \tilde{\mathbf{x}}_\mathfrak{h}) \text{ is true} \}
    \end{aligned}
\end{equation}
where $\kappa''(\tilde{\mathbf{x}}_l,
\tilde{\mathbf{x}}_\mathfrak{h})$ is a slightly modified version of
\eqref{eq: kappa'}, which utilizes the actual scheduled departure time
of load $l$ from $o$, as well as the actual scheduled departure time of
the consolidated load from $h$, instead of the sort level
${dep}(o^\eta_l)$ and ${arr}(o^\eta_{\mathfrak{h}})$, when checking
the time-feasibility at the operational level.

For each $l \in L_C$, let $q_l$ denote the planned volume and $Q_l$
denote the total volume of its trailer type. Define $f^p_l$ as the
cost of using the trailer of load $l$ on path $p$. If $p$ is a direct
route, $f^p_l$ represents the transportation cost of the trailer of
load $l$ from $o_l$ to $d$. If $p$ is a consolidation route, $f^p_l$
instead accounts for the transportation cost of the trailer of load
$l$ from the consolidation point to $d$.  Additionally, let $c^p_{l}$
be the detour cost of transporting load $l$ to the consolidation point
of path $p$. If $p$ is a direct route, and thus $l$ in this case is
not transported to a consolidation point, then $c^p_{l} = 0$.  The
optimization model also needs to reason about trailers and capacities.
For that purpose, it is important to introduce some notations for the
origin of the last leg of path $p$. If $p$ is a direct route $(o,d)$,
then $oll_p = o$.  If $p$ is a consolidation route, then $(o,h,d)$, $oll_p =
h$.

\subsubsection{Optimization Model}\label{sec: consolidation formulations}



For each $l$ and every associated path $p \in P_l$, the optimization model introduces two binary decision variables:
\begin{itemize}
    \item $\xi^p_{l} \in \{0,1\}$ represents whether load $l$ is
      assigned to path $p$;
    \item $\nu^p_{l} \in \{0,1\}$ represents whether the trailer of
      $l$ is used for transport on path $p$. If $\nu^p_{l} = 0$, $l$'s
      trailer is eliminated and $l$ is consolidated into another
      load's trailer.
\end{itemize}

\noindent
The partial loads consolidation optimization problem is formulated as follows: \\
\begin{subequations}
    \begin{align}
        & \min \sum_{l \in L} \sum_{p \in P_l} \left( c^p_{l} \xi^p_{l} + f^p_{l} \nu^p_{l} \right) \label{eq: obj} \\
        \text{s.t.} & \sum_{p \in P_{l}} \xi^p_{l} = 1, \quad \forall \ l \in L \label{eq: constraint select one route} \\
        & \sum_{l \in L} \ \sum_{p \in P_l | oll_p = h} q_{l} \ \xi^p_{l} \leq \sum_{l \in L} \ \sum_{p \in P_l | oll_p = h} Q_{l} \ \nu^p_{l}, \quad \forall \ h \in H \label{eq: constraint capactiy} \\
        & \nu^p_{l} \leq \xi^p_{l}, \quad \forall \ l \in L, \forall \ p \in P_{l} \label{eq: load compatibility} \\
        & \xi^p_{l}, \nu^p_{l} \in \{0, 1\}, \quad \forall \ l \in L, \forall \ p \in P_{l} \label{eq: decision variables}
    \end{align}
    \label{eq: optimization model}
\end{subequations}

\noindent
Constraints \eqref{eq: constraint select one route} ensure that each
load selects exactly one path. Capacity constraints \eqref{eq:
  constraint capactiy} guarantee that sufficient trailers are
available to transport all consolidated loads at $h$. Finally,
compatibility constraints \eqref{eq: load compatibility} ensure that
load $l$ can only provide capacity along path $p$ if load $l$ is
routed along path $p$.  The objective is to minimize the total
transportation and trailer usage costs. Conceptually, this model can
be viewed as a generalized assignment problem featuring variable
capacity constraints, akin to the network design formulations in
\cite{greening-2023, greening-2023-1}.

%% file: method-opt-vis.tex
\subsection{An Illustration of the Optimization Component}

\begin{figure}[!t]

    \centering
    \begin{subfigure}{0.3\linewidth}
        \centering
        \includegraphics[width=\textwidth]{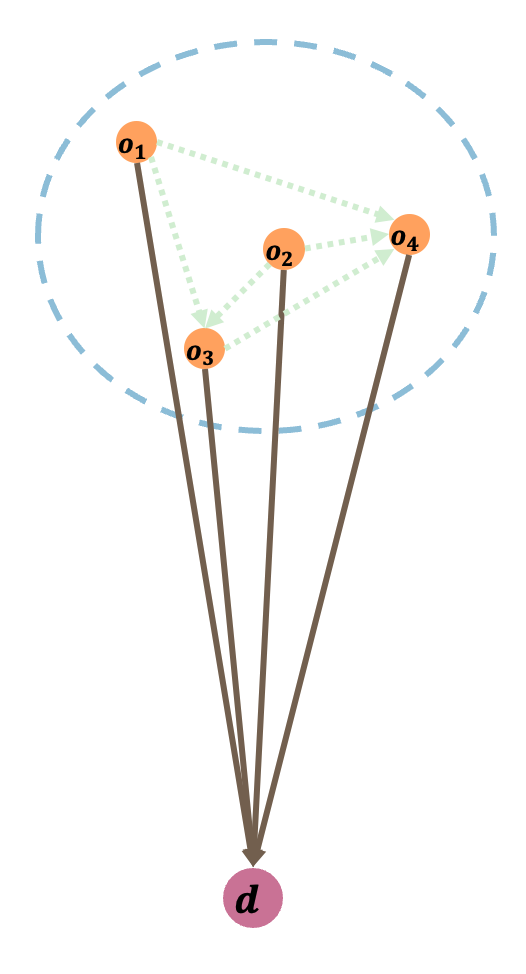}
        \caption{Feasible Path Generation.}
        \label{fig: opt module - feasible path generation}
    \end{subfigure}
    \hfill
    \begin{subfigure}{0.3\linewidth}
        \centering
        \includegraphics[width=\textwidth]{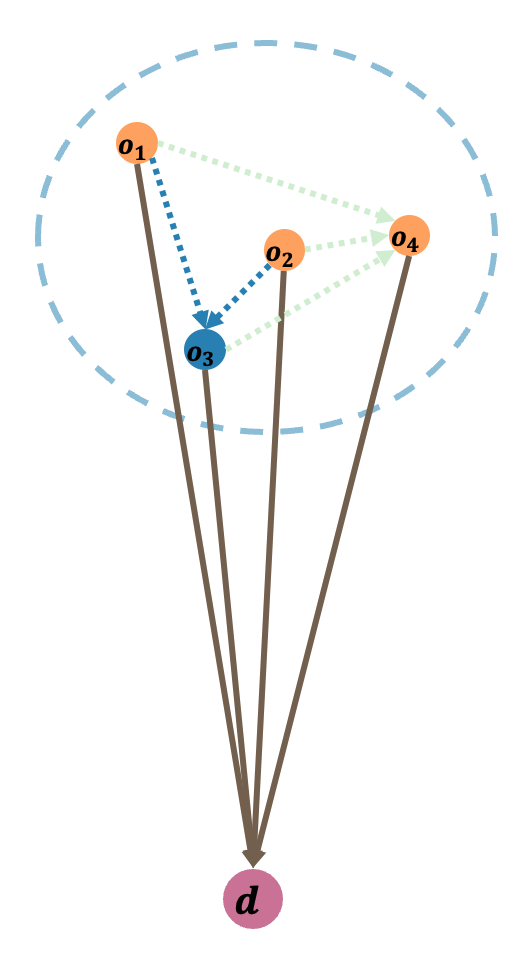}
        \caption{Consolidation Paths Selection.}
        \label{fig: opt module - consolidation path selection}
    \end{subfigure}
    \hfill
    \begin{subfigure}{0.3\linewidth}
        \centering
        \includegraphics[width=\textwidth]{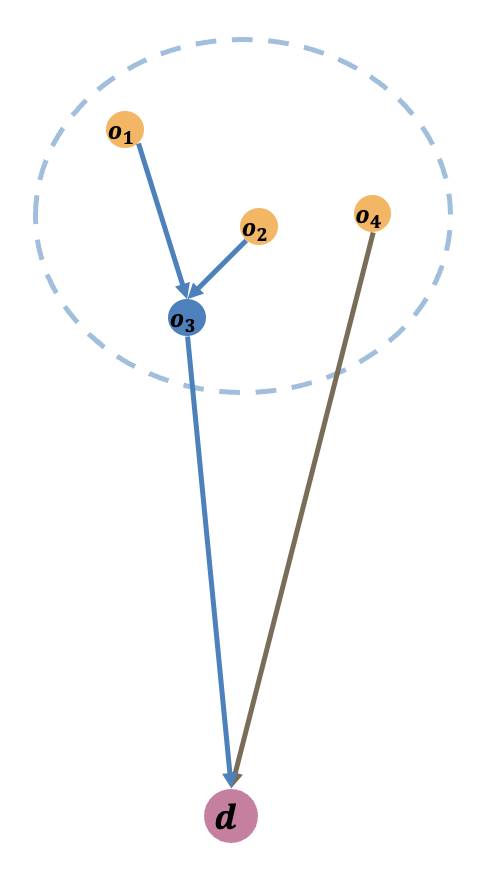}
        \caption{Final Consolidation Results}
        \label{fig: opt module - final consolidation results}
    \end{subfigure}

    \caption{An Illustration of the Optimization Component.}
    \label{fig: an illustration of the OPT module}
    
\end{figure}

Figure~\ref{fig: an illustration of the OPT module} presents an
example to illustrate each step of the optimization component. Consider four
partial loads $l_1, \ldots, l_4 \in L_C$. The corresponding origins
$o_1, \ldots, o_4$ have scheduled departure times $t_1 < t_2 < t_3 < t_4$,
where $o_3$ and $o_4 \in H$ represent consolidation points identified
by the ML component. In Figure~\ref{fig: opt module - feasible path
generation}, dark brown arrows show scheduled load transportation from
$o_1, \ldots, o_4$ to the common destination $d$, and light dashed green
arrows indicate alternative consolidation paths as described in
Section~\ref{sec: feasible path generation}. Then, Figures~\ref{fig:
opt module - consolidation path selection} and~\ref{fig: opt module -
final consolidation results} depict the consolidation decisions made
by the optimization in accordance with Section~\ref{sec: consolidation
formulations}. In Figure~\ref{fig: opt module - consolidation path
selection}, the paths $o_1 \rightarrow o_3$ and $o_2 \rightarrow o_3$
are selected. As shown in Figure~\ref{fig: opt module - final
consolidation results}, the three originally scheduled trailers from
$o_1$, $o_2$, and $o_3$ to $d$ are reduced to two better-utilized
trailers due to consolidation at $o_3$, resulting in fewer trailers,
less total travel distance, and lower overall transportation costs.

%% file: experiments.tex
\section{Experimental Setting}
\label{sec: experiments}

\spot{} was evaluated through extensive experiments conducted on a
real-world freight transportation dataset provided by the industrial
partner. The experiments evaluate the performance of \spot{} against
the non-consolidation TruckLoad (TL) load planning currently employed
by the industrial partner. They also consider the improvements
compared to a nearest-neighbor-based heuristic (NNCH) algorithm for
load consolidation. The section describes the experimental setting,
including an overview of the dataset, the baselines, and the comparison metrics.

\subsection{Datasets}

The dataset comprises six months of freight transportation records
provided by the industry partner. It contains over two million loads,
39\% of which are categorized as partial loads, defined as having a
capacity utilization below 80\%. The records cover approximately one
thousand terminals throughout the United States and include thousands
of terminal-and-sort combinations serving as load origins and
destinations.

\begin{figure}[!t]

    \centering

    \begin{subfigure}{0.45\linewidth}
        \centering
        \includegraphics[width=\textwidth]{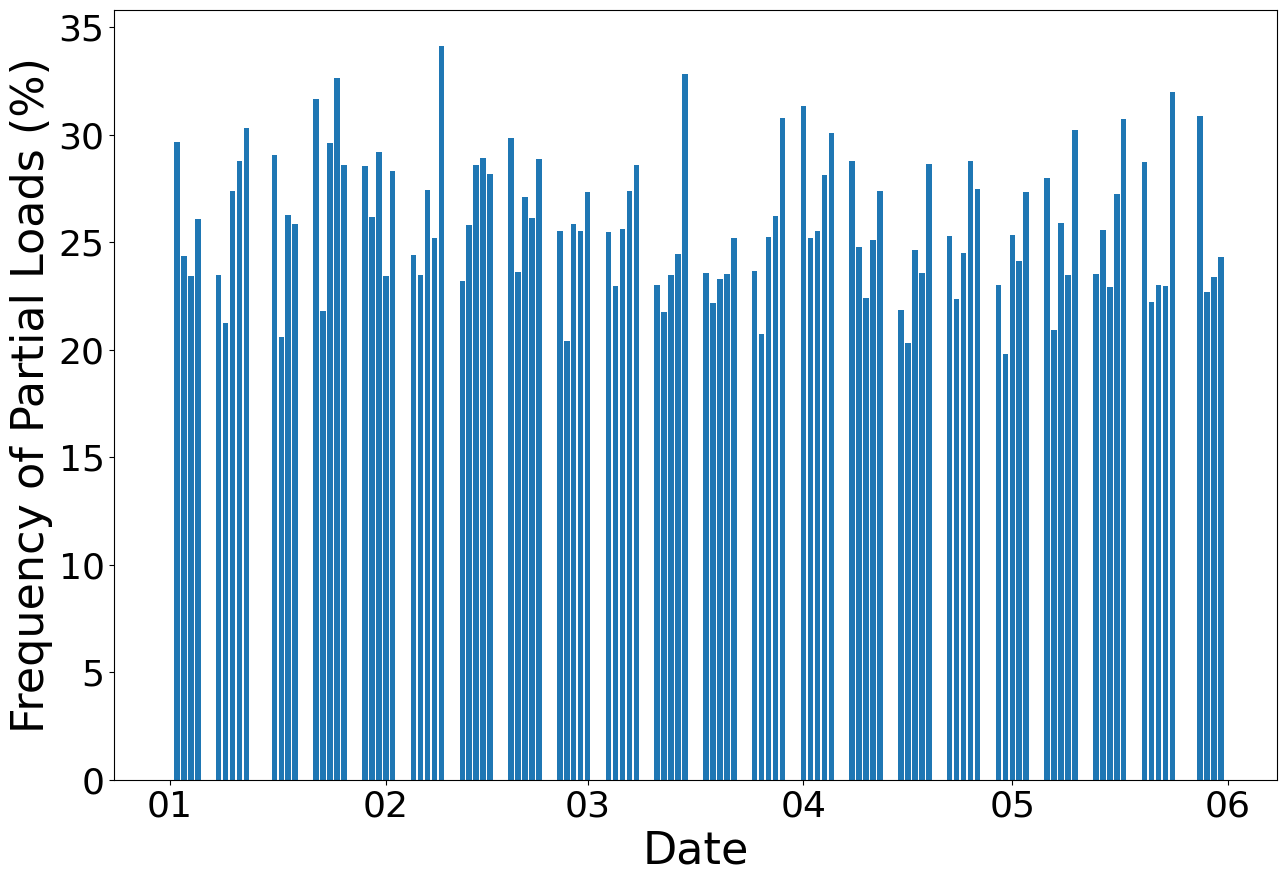}
        \caption{Daily Count of Partial Loads.}
        \label{fig: partial load freq}
    \end{subfigure}
    \hfill
    \begin{subfigure}{0.45\linewidth}
        \centering
        \includegraphics[width=\textwidth]{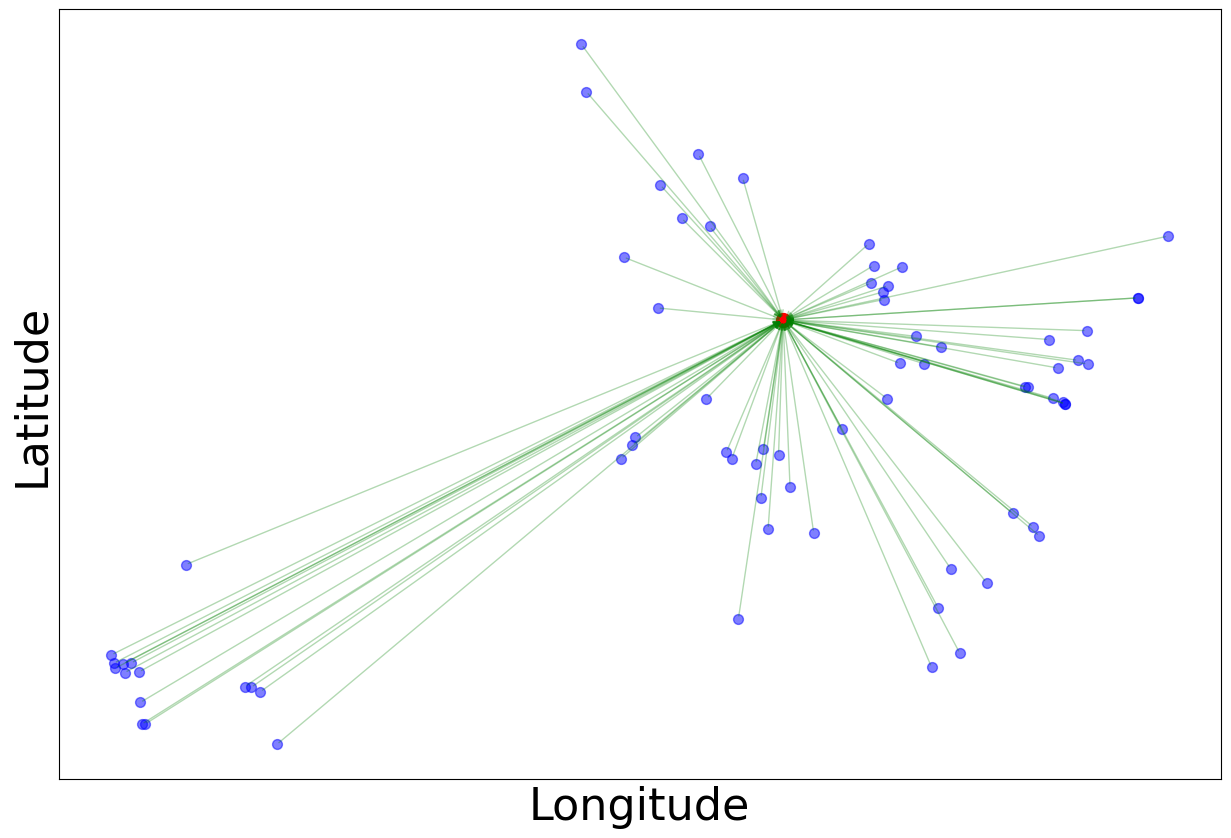}
        \caption{Spatial Distribution of Partial Loads.}
        \label{fig: partial load spatial}
    \end{subfigure}

    \caption{Daily Count and Spatial Distribution of Partial Loads for a Specific Destination.}
    \label{fig: basic data analysis}
    
\end{figure}

A basic analysis and visualization of historical data revealed
potential consolidation opportunities. Figure~\ref{fig: partial load
  freq} shows that, for a specific destination, partial loads occur consistently
on weekdays, with the daily percentage of partial loads exceeding 25\%
on average. Moreover, for the same destination on a specific
operational day, the spatial distribution of partial loads exhibits
clustering characteristics, which can be used to facilitate load
consolidation (Figure ~\ref{fig: partial load spatial}).

In the six-month load dataset, the load data from the final three
weeks was utilized as the testing dataset for algorithm
comparison. The remaining data was used as training input for the ML
component of \spot{}. Thirty destinations are selected for experiments
exhibiting varying numbers of daily partial load occurrences. Among
these, five destinations (\textbf{Tier 1}) had the highest daily
volume of partial loads, ten destinations (\textbf{Tier 2}) had a
moderate daily volume, and the remaining fifteen destinations
(\textbf{Tier 3}) had relatively low levels of daily partial
loads. Each destination is evaluated over 15 consecutive weekdays
(3 weeks), and the average performance is reported.

\subsection{Baselines}

\spot{} is compared against two different baselines. 

\textit{TruckLoad (TL):} The TL transportation load planning approach
mirrors the current strategy employed by the industrial partner and
does not consider load consolidation options. Regardless of its
utilization level, each load is directly transported from origin to
destination. TL is a baseline in almost every study within this domain
\cite{oguntola-2023, aboutalib-2024, baykasoglu-2011,
  mesa-arango-2013, monsreal-2024, kay-2021}.

\textit{Nearest Neighbor Consolidation Heuristic (NNCH) Algorithm:}
The NNCH algorithm is introduced as an enhancement of TL, to show the
benefits of the \spot{} optimization component over a simple informed
heuristics. NNCH follows the general principles of the
nearest-neighbor heuristic proposed by \citet{monsreal-2024}, with
necessary modifications to adapt to terminal consolidation. This
adaptation is essential because NNCH was initially designed for the
m-PD-VRPTW formulation. Algorithm~\ref{alg:nnch_alg} describes the
adapted NNCH heuristic which operates as follows: for each load, the
algorithm iterates through all other loads, starting from the nearest
and proceeding to the farthest, and consolidates loads as long as the
capacity constraints are satisfied. The input to NNCH matches the
input used in \spot’s consolidation optimization, i.e., the set of
loads $L_C$.

\begin{algorithm}[!t]
\caption{Nearest Neighbor Consolidation Heuristic (NNCH)}
\label{alg:nnch_alg}
\SetAlgoLined
\KwIn{Set of all feasible paths ($\textbf{paths}$), where each path contains origin information $(o, o\_quantity, o\_capacity, o\_departure\_time)$, intermediate terminal information  $(h, h\_quantity, h\_capacity)$, and travel time between them $o\_h\_travel\_time$.}
\KwOut{List of consolidation decisions $(o, h)$.}
Sort $\textbf{paths}$ by $o\_departure\_time$\;
Initialize $consolidated\_decisions \gets []$\;
\While{$\textbf{paths}$ is not empty}{
    Extract the first path and corresponding $\mathbf{o}$\;
    Find candidates $C \subseteq \textbf{paths}$ where $o = \mathbf{o}$ and $h \neq \mathbf{o}$\;
    $consolidated \gets False$\;
    Sort $C$ by $o\_h\_travel\_time$ (ascending)\;
    \ForEach{$candidate \in C$}{
        Extract $(h, h\_quantity, h\_capacity)$ from candidate\;
        \If{$o\_quantity + h\_quantity \leq \max \{ o\_capacity, h\_capacity \}$}{
            Add $(\mathbf{o}, h)$ to $consolidated\_decisions$\;
            Remove paths involving $o$ and $h$ from $\textbf{paths}$\;
            $consolidated \gets True$\;
            \textbf{break}\;
        }
    }
    \If{\textbf{not} $consolidated$}{
        Add $(\mathbf{o}, \mathbf{o})$ to $consolidated\_decisions$ (fallback to TL)\;
        Remove paths involving $\mathbf{o}$ from $\textbf{paths}$\;
    }
}
\Return $consolidated\_decisions$\;
\end{algorithm}

\subsection{Metrics}

Two types of evaluation are performed to understand the contribution
pf \spot{} and its components. The first type of evaluation analyzes
the effectiveness of the operational consolidation decisions by
measuring normalized total travel distance (\textit{Travel Distance
  (\%)}), the percentage reduction in transportation costs
(\textit{Cost Reduction (\%)}), and the percentage of partial loads
cut (\textit{Loads Cut (\%)}) for selected operational days and
destinations. These metrics enable a comparative analysis of \spot{}
against NNCH and TL. The second type of evaluation measures the
contributions of the ML component. For consolidation points, the
evaluation reports: \textit{Coverage}, the ratio of consolidated partial
loads to the total number of partial loads; \textit{CP Ratio}, the
proportion of partial load origins serving as consolidation points;
and \textit{Daily Loads Per CP}, the average number of loads processed
per consolidation point. The evaluation reports route-related metrics
including \textit{Path Freq.} that captures the frequency of optimal
consolidation routes linked to consolidation points and \textit{Num of
  Paths} that captures the proportion of feasible routes selected by
\spot{}. Together, these metrics highlight the effectiveness of the ML
component in identifying recurrent consolidation opportunities,
narrowing the optimization search space, enhancing consolidation
decisions, and offering insights for tactical load planning.

%% file: experimental-result.tex
\section{Experimental Results}
\label{sec: experiment-result}

This section presents the experimental results of \spot{}. The primary
goal is to address the following research questions: \textit{(Q1)
  Operational-Level Performance:} Can \spot{} provide effective
consolidation decisions at an operational level, directly enabling
cost reductions in industrial settings?  \textit{(Q2) Tactical
  Insights from Historical Data:} Can \spot{} extract meaningful
information from historical data that supports load consolidation
decisions and offers insights for tactical planning?  \textit{(Q3)
  Computational Efficieny:} Is \spot{} sufficient to handle load
consolidation across a large-scale transportation network, or is its
performance constrained by computational complexity?

\subsection{Consolidation Performance}

\begin{figure}[!t]

    \centering

    \begin{subfigure}{0.3\linewidth}
        \centering

        \includegraphics[width=\textwidth]{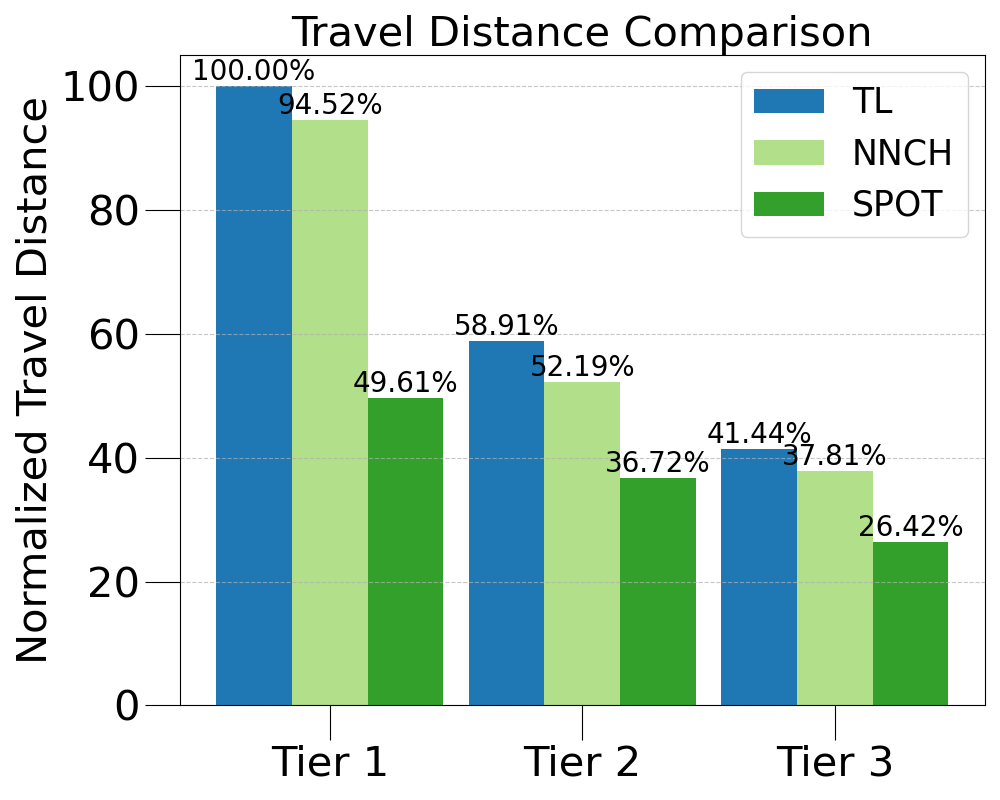}
        \caption{Travel Distance}
        \label{fig: Travel Distance Comparison by Tier}
    \end{subfigure}
    \hspace{0.01\linewidth}  
    \begin{subfigure}{0.3\linewidth}
        \centering
        \includegraphics[width=\textwidth]{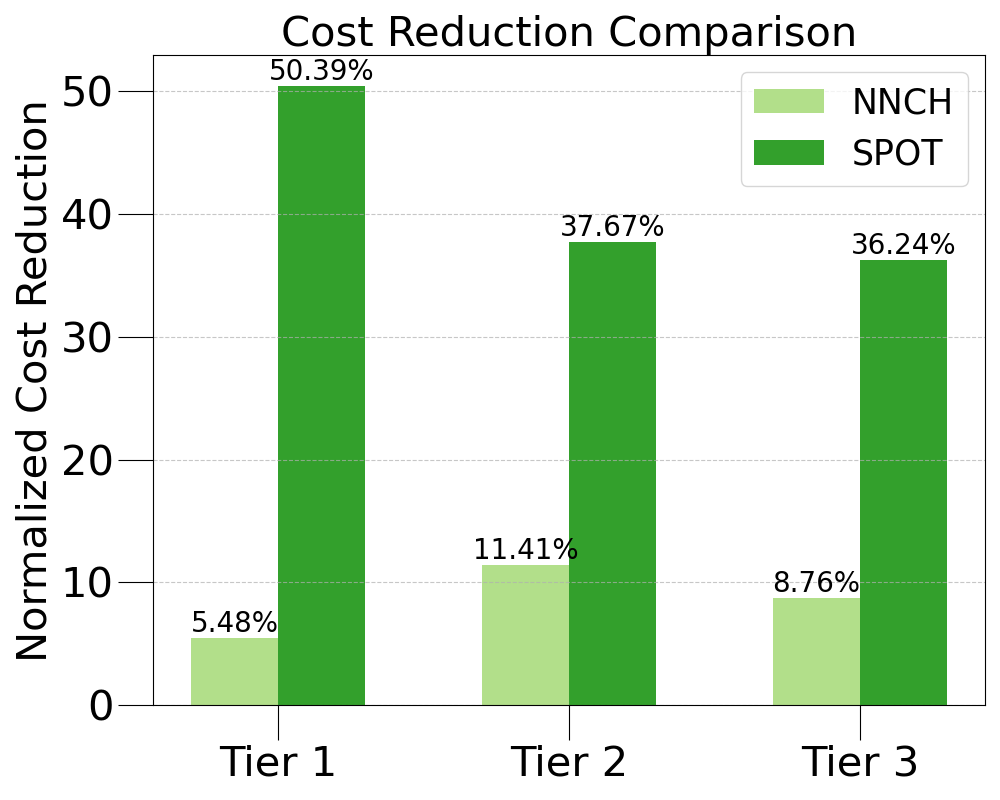}
        \caption{Cost Reduction}
        \label{fig: Cost Reduction Comparison by Tier}
    \end{subfigure}
    \hspace{0.01\linewidth}  
    \begin{subfigure}{0.3\linewidth}
        \centering
        \includegraphics[width=\textwidth]{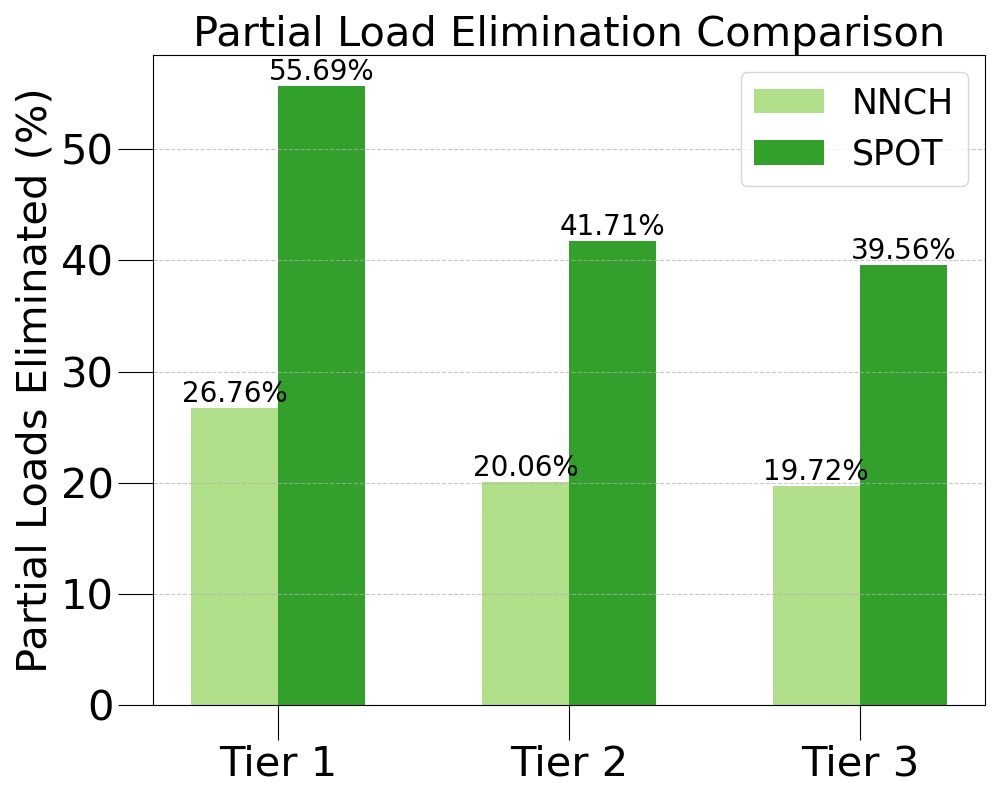}
        \caption{The Number of Partial Loads Cut}
        \label{fig: num load cuts by Tier}
    \end{subfigure}

    \caption{Consolidation Performance Comparison by Tier.}
    \label{fig: operational level performance}
    
\end{figure}

Let $\mathbf{\epsilon}$ denote the maximum threshold for grouping
nodes in the clustering phase, and $\mathbf{\tau}$ the minimum
frequency of patterns in historical data for CFIM. Figures~\ref{fig:
  operational level performance} and \ref{fig: consolidation points
  statistics} present results for the configuration
$(\mathbf{\epsilon}=0.30, \mathbf{\tau}=5)$, which represents the
optimal setting for \spot{}. Additional experimental results under
various $(\mathbf{\epsilon}, \mathbf{\tau})$ combinations are provided
in Tables~\ref{table: td-cr-partial-loads} and \ref{table: cp
  statistics}. Empirically, $\mathbf{\epsilon} \in \{0.2, 0.25, 0.3\}$
corresponds to clustering angles of approximately $20^\circ \sim
30^\circ$, while $\mathbf{\tau} \in \{5, 10\}$ captures patterns
occurring at least monthly or biweekly.

Figure \ref{fig: operational level performance} presents a comparison
of \spot’s consolidation performance against TL and NNCH across the
three destination tiers. By combining the ML component with feasible
path generation and optimization, \spot{} delivers efficient
consolidation outcomes. It consistently outperforms the alternatives
across all tiers.

Compared to the existing operational method (TL), \spot{} delivers
notable reductions in both travel distance and transportation
costs. For Tier-1 destinations, which handle high daily volumes of
partial loads, \spot{} improves performance compared to TL by roughly
50\%.  Even in Tier-2 and Tier-3 destinations, where consolidation
opportunities are limited due to lower volumes, it maintains cost
savings of over 36\% (Figure~\ref{fig: Travel Distance Comparison by
  Tier}). \spot{} achieves cost reductions of at least three times
those of NNCH, in Tier-2 and up to nine times in Tier-1
(Figure~\ref{fig: Cost Reduction Comparison by Tier}). In terms of the
number of loads being cut, the difference between NNCH and \spot{} in NLC is
relatively small compared to their large gap in cost reduction. Across
all three tiers, NNCH cuts about half as many loads as \spot{} with
significantly lower cost reductions, especially in Tier-1, where the
savings are far below half of what \spot{} achieves. This highlights
NNCH’s shortcoming as a greedy heuristic, often making poor
consolidation choices. This pattern is further illustrated in
Table~\ref{table: td-cr-partial-loads}, where the best results for TL,
NNCH, and \spot{} are marked in bold. Interestingly, NNCH’s best cost
reductions do not align with its highest number of loads cut. In
contrast, \spot{} consistently shows that smarter consolidation leads
to a clear, positive correlation between the number of loads cut and
cost savings. The superior consolidation performance of \spot{}
provides a strong answer to \textit{\textbf{(Q1)}}, demonstrating its
ability to deliver the best operational-level performance.

\begin{table}[!t]
\centering
\begin{tabular}{c|cccccc}
\hline
\textbf{} & $\boldsymbol{\epsilon}$ & $\boldsymbol{\tau}$ & \textbf{Method} & \textbf{Travel Distance (\%)} & \textbf{Cost Reduction(\%)} & \textbf{Loads Cut (\%)} \\ \hline

\multirow{12}{*}{\rotatebox{90}{\textbf{Tier 1}}}
& - & - & \textbf{TL} & \textbf{100.0} & - & - \\ \cline{2-7}

& \multirow{4}{*}{0.20}
  & \multirow{2}{*}{5}
    & NNCH & 94.07 & 5.93 & 25.38 \\
& & & SPOT & 50.38 & 49.62 & 54.94 \\ \cline{3-7}
& & \multirow{2}{*}{10}
    & \textbf{NNCH} & \textbf{90.08} & \textbf{9.92} & \textbf{23.09} \\
& & & SPOT & 54.19 & 45.81 & 51.01\\ \cline{2-7}

& \multirow{4}{*}{0.25}
  & \multirow{2}{*}{5}
    & NNCH & 94.06 & 5.94 & 26.01 \\
& & & SPOT & 50.25 & 49.75 & 55.28 \\ \cline{3-7}
& & \multirow{2}{*}{10}
    & NNCH & 90.53 & 9.47 & 24.05 \\
& & & SPOT & 53.55 & 46.45 & 51.59 \\ \cline{2-7}

& \multirow{4}{*}{0.30}
  & \multirow{2}{*}{5}
    & NNCH & 94.52 & 5.48 & 26.76 \\
& & & \textbf{SPOT} & \textbf{49.61} & \textbf{50.39} & \textbf{55.69} \\ \cline{3-7}
& & \multirow{2}{*}{10}
    & NNCH & 91.25 & 8.75 & 25.0 \\
& & & SPOT & 52.97 & 47.03 & 52.45 \\ \hline

\multirow{12}{*}{\rotatebox{90}{\textbf{Tier 2}}}
& - & - & \textbf{TL} & \textbf{58.91} & - & - \\ \cline{2-7}

& \multirow{4}{*}{0.20}
  & \multirow{2}{*}{5}
    & NNCH & 51.66 & 12.31 & 18.06 \\
& & & SPOT & 38.33 & 34.94 & 38.82 \\ \cline{3-7}
& & \multirow{2}{*}{10}
    & \textbf{NNCH} & \textbf{48.82} & \textbf{17.13} & \textbf{15.66} \\
& & & SPOT & 41.06 & 30.3 & 33.77 \\ \cline{2-7}

& \multirow{4}{*}{0.25}
  & \multirow{2}{*}{5}
    & NNCH & 52.19 & 11.41 & 19.19 \\
& & & SPOT & 37.47 & 36.4 & 40.38 \\ \cline{3-7}
& & \multirow{2}{*}{10}
    & NNCH & 49.21 & 16.47 & 16.67 \\
& & & SPOT & 39.99 & 32.12 & 35.48 \\ \cline{2-7}

& \multirow{4}{*}{0.30}
  & \multirow{2}{*}{5}
    & NNCH & 52.19 & 11.41 & 20.06 \\
& & & \textbf{SPOT} & \textbf{36.72} & \textbf{37.67} & \textbf{41.71} \\ \cline{3-7}
& & \multirow{2}{*}{10}
    & NNCH & 49.56 & 15.88 & 17.7 \\
& & & SPOT & 39.16 & 33.53 & 37.09 \\ \hline

\multirow{12}{*}{\rotatebox{90}{\textbf{Tier 3}}}
& - & - & \textbf{TL} & \textbf{41.44} & - & - \\ \cline{2-7}

& \multirow{4}{*}{0.20}
  & \multirow{2}{*}{5}
    & NNCH & 36.75 & 11.32 & 17.25 \\
& & & SPOT & 27.36 & 33.98 & 37.14 \\ \cline{3-7}
& & \multirow{2}{*}{10}
    & \textbf{NNCH} & \textbf{34.06} & \textbf{17.82} & \textbf{15.43} \\
& & & SPOT & 28.91 & 30.24 & 32.56 \\ \cline{2-7}

& \multirow{4}{*}{0.25}
  & \multirow{2}{*}{5}
    & NNCH & 37.69 & 9.06 & 18.86 \\
& & & SPOT & 26.73 & 35.5 & 38.51 \\ \cline{3-7}
& & \multirow{2}{*}{10}
    & NNCH & 35.0 & 15.54 & 16.75 \\
& & & SPOT & 28.37 & 31.54 & 34.07 \\ \cline{2-7}

& \multirow{4}{*}{0.30}
  & \multirow{2}{*}{5}
    & NNCH & 37.81 & 8.76 & 19.72 \\
& & & \textbf{SPOT} & \textbf{26.42} & \textbf{36.24} & \textbf{39.56} \\ \cline{3-7}
& & \multirow{2}{*}{10}
    & NNCH & 35.3 & 14.8 & 17.34 \\
& & & SPOT & 28.14 & 32.09 & 34.78 \\ \hline
\end{tabular}
\vspace{0.2cm}
\caption{Travel Distance, Cost Reduction, and the Number of Partial Loads Cut.}
\label{table: td-cr-partial-loads}
\end{table}

\subsection{Statistics regarding Consolidation Points}

\begin{figure}[!t]

    \centering

    \begin{subfigure}{0.3\linewidth}
        \centering
        \includegraphics[width=\textwidth]{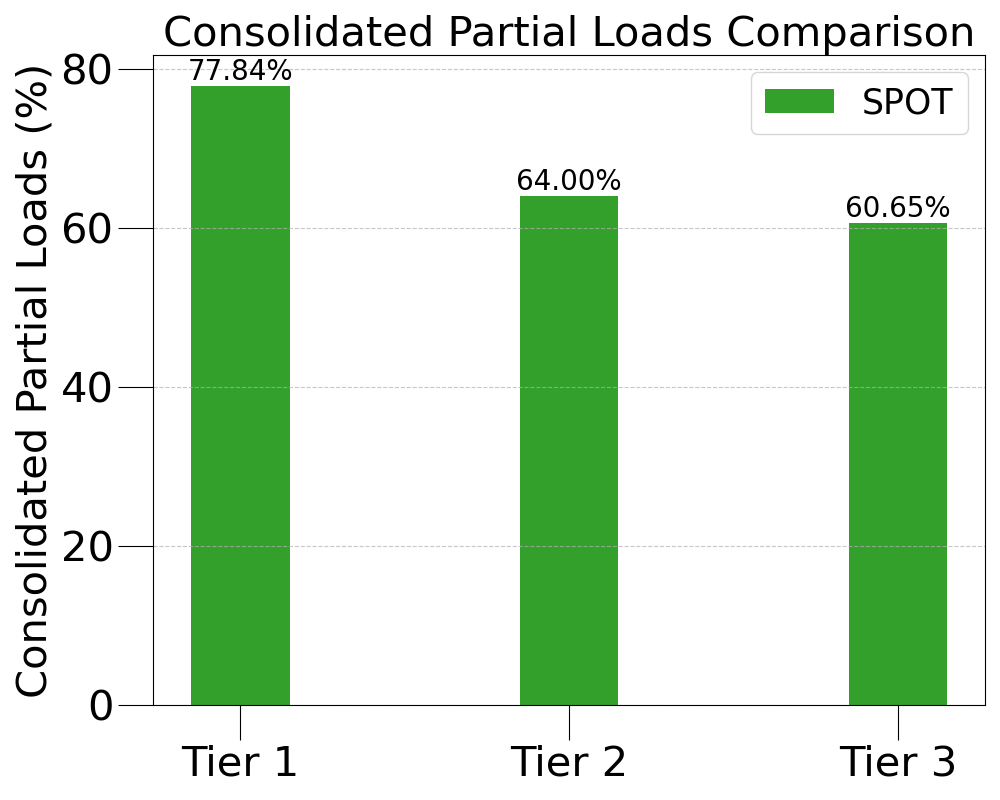}
        \caption{Cov}
        \label{fig: Coverage}
    \end{subfigure}
    \hspace{0.01\linewidth}  
    \begin{subfigure}{0.3\linewidth}
        \centering
        \includegraphics[width=\textwidth]{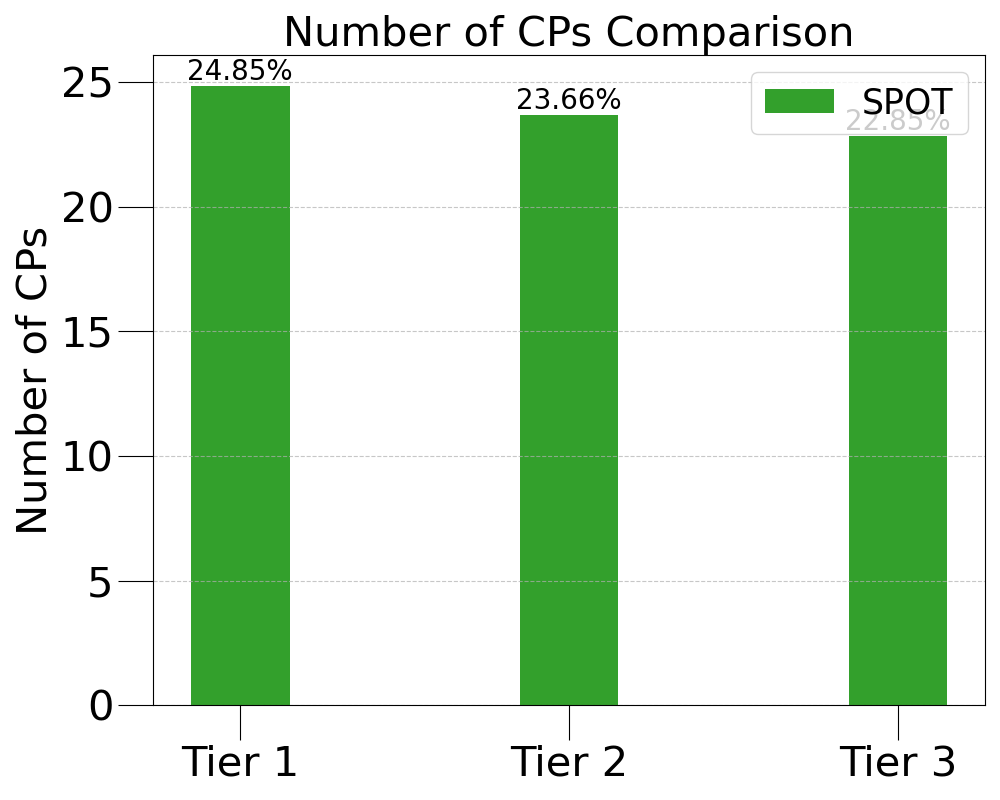}
        \caption{CP Ratio}
        \label{fig: CP Ratio}
    \end{subfigure}
    \hspace{0.01\linewidth}  
    \begin{subfigure}{0.3\linewidth}
        \centering
        \includegraphics[width=\textwidth]{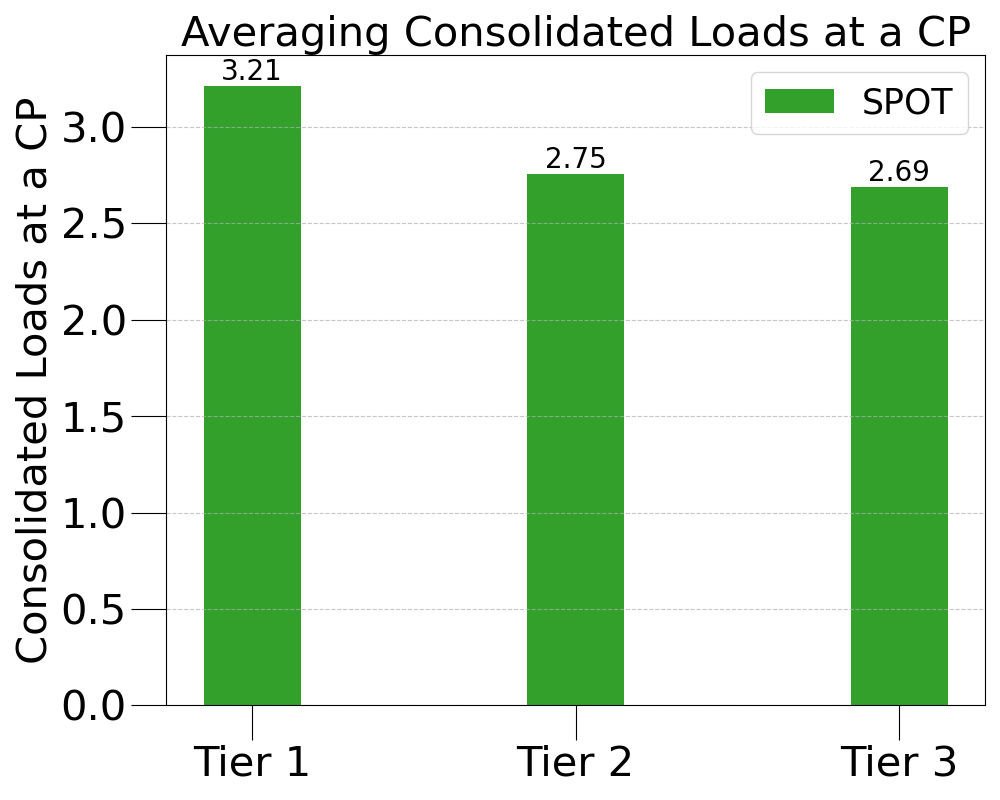}
        \caption{Loads Per CP}
        \label{fig: Loads Per CP}
    \end{subfigure}

    \caption{Consolidation Points Statistics}
    \label{fig: consolidation points statistics}
    
\end{figure}

Figure~\ref{fig: consolidation points statistics} demonstrates the
strong synergy between the ML and optimization components. When using
\spot{}, 60\%–80\% of daily partial loads are consolidated at just
20\% of origins, which act as consolidation points. This low ratio of
consolidation points significantly eases tactical-level planning for
the industry, as fewer terminals require adjustments, making
operational execution of load consolidation more practical. These
efforts are well justified, given that most daily partial loads are
included in the consolidation process. Besides, as shown in
Table~\ref{table: cp statistics} -- \textit{Num of Paths}, \spot{}
considers only around 40\%–50\% of all time-feasible consolidation
paths. Importantly, these selected routes are those that frequently
appear in historical records (Table~\ref{table: cp statistics} --
\textit{Path Freq}). These facts demonstrate the ML component’s
ability to recognize repeating patterns and significantly reduce the
search space for optimization.

Furthermore, Tables~\ref{table: td-cr-partial-loads} and \ref{table:
  cp statistics} highlight a notable trend in the behavior of the ML
module, which offers valuable insights for tactical
planning. Specifically, increasing $\mathbf{\epsilon}$ (loosening
clustering constraints) or decreasing $\mathbf{\tau}$ (lowering the
threshold for frequent patterns), or both, leads to greater cost
reduction (\textit{Cost Reduction} in Table~\ref{table: cp
  statistics}) and an increase in \textit{CP Ratio} and \textit{Num of
  Paths} in Table~\ref{table: cp statistics}. As discussed previously,
from an industry perspective, it is crucial to undertake
tactical-level preparations by coordinating the load consolidation
planning with other components of the transportation management
system, such as driver scheduling, route assignments, equipment
availability, and intermediate terminal arrangements. In the
experiments, the associated preparation costs are proportional to
\textit{CP Ratio}, the number of paths requiring modifications or
adjustments at the tactical level to ensure that the chosen
consolidation routes are operationally feasible and achieve cost
reductions. The associated preparation costs increase with the
\textit{CP Ratio} and the \textit{Num of Paths}, as both metrics
reflect how many terminal and route candidates may require tactical
adjustments to ensure that the selected consolidation plans are
feasible in practice. Consequently, \spot{} provides decision makers
with a clear trade-off between the operational cost of advanced effort
(i.e., how many candidates to prepare for) and the potential
transportation cost savings. Decision makers can make informed choices
by examining the exact paths identified by the ML component and their
corresponding performance at both the tactical and operational levels.

\begin{table}[!t]
\centering
\begin{tabular}{c|cccccccc}
\hline
\textbf{} & $\boldsymbol{\epsilon}$ & $\boldsymbol{\tau}$ & \textbf{Path Freq (\%)} & \textbf{Num of Paths (\%)} & \textbf{Coverag (\%)} & \textbf{CP Ratio (\%)} & \textbf{Daily Loads Per CP} \\ \hline

\multirow{6}{*}{\rotatebox{90}{\textbf{Tier 1}}}
& \multirow{2}{*}{0.20} & 5 & 30.25 & 47.74 & 76.54 & 24.34 & 3.22 \\
& & 10 & 31.8 & 39.19 & 71.32 & 23.05 & 3.2 \\ \cline{2-8}
& \multirow{2}{*}{0.25} & 5 & 30.49 & 50.71 & 76.8 & 24.37 & 3.23 \\
& & 10 & 31.94 & 43.47 & 71.99 & 23.17 & 3.19 \\ \cline{2-8}
& \multirow{2}{*}{0.30} & 5 & 30.38 & 53.66 & 77.84 & 24.85 & 3.21 \\
& & 10 & 31.87 & 46.7 & 73.01 & 23.41 & 3.22 \\ \hline

\multirow{6}{*}{\rotatebox{90}{\textbf{Tier 2}}}
& \multirow{2}{*}{0.20} & 5 & 25.11 & 35.18 & 60.15 & 22.63 & 2.7 \\
& & 10 & 27.27 & 29.17 & 53.05 & 20.48 & 2.62 \\ \cline{2-8}
& \multirow{2}{*}{0.25} 
  & 5 & 25.22 & 38.09 & 62.19 & 23.16 & 2.72 \\
& & 10 & 27.07 & 31.41 & 55.63 & 21.15 & 2.66 \\ \cline{2-8}
& \multirow{2}{*}{0.30} 
  & 5 & 25.36 & 41.95 & 64.0 & 23.66 & 2.75 \\
& & 10 & 27.14 & 34.64 & 57.68 & 21.71 & 2.71 \\ \hline

\multirow{6}{*}{\rotatebox{90}{\textbf{Tier 3}}}
& \multirow{2}{*}{0.20}
  &  5 & 23.97 & 40.61 & 56.67 & 21.49 & 2.68 \\
& & 10 & 26.23 & 34.7 & 50.7 & 19.56 & 2.61 \\ \cline{2-8}
& \multirow{2}{*}{0.25}
  &  5 & 24.2 & 44.94 & 59.21 & 22.42 & 2.67 \\
& & 10 & 26.56 & 38.17 & 53.16 & 20.6 & 2.59 \\ \cline{2-8}
& \multirow{2}{*}{0.30} & 5 & 24.34 & 47.43 & 60.65 & 22.85 & 2.69 \\
& & 10 & 26.66 & 39.96 & 54.23 & 20.91 & 2.61 \\ \hline
\end{tabular}
\vspace{0,2cm}
\caption{Consolidation Points \& Paths Statistics of \spot{}.}
\label{table: cp statistics}
\end{table}

In summary, the observed consolidation performance supports the
conclusion that \spot{} effectively extracts critical insights from
historical data at the tactical level and can develop effective,
detailed consolidation decisions at the operational level, thereby
addressing questions \textit{Q1} and \textit{Q2}.

\subsection{Computational Efficieny}

\spot{} considers each destination independently, offering two primary
advantages when scaling to the entire transportation network. First,
parallelizing the computations across multiple destinations is
straightforward because each destination is computed independently,
resulting in high overall efficiency. Second, restricting the
destination-based consolidation leads to a binary optimization model
that can be solved to optimality quickly. For Tier 1 destinations,
\spot{} requires an average of 18 seconds of computation time while,
for Tiers 2 and 3, it only requires 5 seconds and 2 seconds,
respectively, demonstrating its ability to handle network-wide
operational scenarios efficiently. These results demonstrate \spot’s
capacity to efficiently handle network-wide operational scenarios,
thereby addressing \textit{Q3} as well.

%% file: conclusion.tex
\section{Conclusion}
\label{sec:conclusion}

This study introduces \spot{}, a novel framework for load
consolidation that integrates machine learning techniques with
optimization to improve load consolidation. The ML component combines
spatio-temporal clustering with constrained frequent itemset mining
(CFIM), while the optimization component employs a MIP model to ensure
feasible and cost-effective decisions. By bridging tactical insights
with operational constraints, \spot{}, not only provides actionable
guidance at the tactical level, but also delivers efficient
consolidation decisions to guide route selection on operational days.
Extensive experiments on real-world load data demonstrate \spot{}’s
effectiveness, showing consistent and substantial cost reductions
compared to baseline methods. \spot{} also serves as a blueprint for
further research on combining ML and optimization models for logistics
and supply chain applications, underscoring the benefits of leveraging
historical data in today’s era of data abundance.